\documentclass{article} 
\usepackage{iclr2027_conference,times}

\usepackage[utf8]{inputenc} 
\usepackage[T1]{fontenc}    
\usepackage{hyperref}       
\usepackage{url}            
\usepackage{booktabs}       
\usepackage{amsfonts}       
\usepackage{nicefrac}       
\usepackage{microtype}      
\usepackage{xcolor}         

\usepackage{natbib}
\usepackage{amsmath,amsthm,amsfonts,amssymb, mathtools,bm}
\usepackage[shortlabels]{enumitem}
\usepackage{subcaption}
\usepackage[capitalize,noabbrev]{cleveref}
\usepackage{tikz}
\usepackage{bbm}
\usepackage{footnote}
\usepackage{multirow}
\usepackage{tcolorbox}
\usepackage{tkz-graph}
\usetikzlibrary{backgrounds}
\usetikzlibrary{shapes.geometric} 
\usetikzlibrary{matrix} 

\usepackage{tkz-graph}
\usepackage{amsmath}
\usetikzlibrary{positioning, decorations.pathreplacing, calc, matrix}
\usetikzlibrary{patterns.meta}

\usepackage{graphicx}
\usepackage{xspace}
\usepackage{caption}
\usepackage{subcaption}
\usepackage{wrapfig}        
\usetikzlibrary{arrows.meta, positioning, shapes.geometric, calc}

\usepackage{makecell}
\usetikzlibrary{shadows}

\usepackage{algorithm}
\usepackage{algpseudocode}

\usepackage[normalem]{ulem}

\hypersetup{
    colorlinks=true,       
    linkcolor=azure,       
    urlcolor=blue,         
    citecolor=azure  
}

\usepackage{amsmath,amsfonts,bm}

\def\eqref#1{equation~\ref{#1}}

\def\1{\bm{1}}

\def\ve{{\bm{e}}}

\def\vs{{\bm{s}}}

\def\vz{{\bm{z}}}

\def\mM{{\bm{M}}}

\def\mX{{\bm{X}}}
\def\mY{{\bm{Y}}}

\DeclareMathAlphabet{\mathsfit}{\encodingdefault}{\sfdefault}{m}{sl}
\SetMathAlphabet{\mathsfit}{bold}{\encodingdefault}{\sfdefault}{bx}{n}

\def\gA{{\mathcal{A}}}

\def\gE{{\mathcal{E}}}

\def\gG{{\mathcal{G}}}

\def\gI{{\mathcal{I}}}

\def\gQ{{\mathcal{Q}}}

\def\gV{{\mathcal{V}}}

\def\sR{{\mathbb{R}}}

\theoremstyle{plain}

\makeatletter
\newcommand\tinysmall{\@setfontsize\tinysmall{8}{10}}
\makeatother

\newcommand{\method}{\textsc{Universal Classifier}\xspace}
\newcommand{\uc}{\textsc{UC}\xspace}

\definecolor{white}          {RGB}{255, 255, 255} 
\definecolor{softYellow}{RGB}{255, 255, 50} 
\definecolor{softGreen}{RGB}{50, 255, 50}  
\definecolor{darkpink}{rgb}{0.91, 0.33, 0.5}
\definecolor{azure}{rgb}{0.0, 0.5, 1.0}
\definecolor{awesomered}{rgb}{1.0, 0.13, 0.32}
\definecolor{mikadoyellow}{rgb}{1.0, 0.77, 0.05}
\definecolor{sapphire}{rgb}{0.03, 0.15, 0.4}
\definecolor{darkred}{rgb}{0.6, 0.0, 0.0}
\definecolor{lightgray}{rgb}{0.83, 0.83, 0.83}
\definecolor{lightblue}{RGB}{135, 206, 250}
\newif\ifcuboidshade
\newif\ifcuboidemphedge

\tikzset{
  cuboid/.is family,
  cuboid,
  shiftx/.initial=0,
  shifty/.initial=0,
  dimx/.initial=3,
  dimy/.initial=3,
  dimz/.initial=3,
  scale/.initial=1,
  densityx/.initial=1,
  densityy/.initial=1,
  densityz/.initial=1,
  rotation/.initial=0,
  anglex/.initial=0,
  angley/.initial=90,
  anglez/.initial=225,
  scalex/.initial=1,
  scaley/.initial=1,
  scalez/.initial=0.5,
  front/.style={draw=black,fill=white},
  top/.style={draw=black,fill=white},
  right/.style={draw=black,fill=white},
  shade/.is if=cuboidshade,
  shadecolordark/.initial=black,
  shadecolorlight/.initial=white,
  shadeopacity/.initial=0.15,
  shadesamples/.initial=16,
  emphedge/.is if=cuboidemphedge,
  emphstyle/.style={thick},
  zcut/.initial=100,
  topcut/.style={draw=black,fill=white},
  rightcut/.style={draw=black,fill=white},
}

\newcommand{\tikzcuboidkey}[1]{\pgfkeysvalueof{/tikz/cuboid/#1}}

\newcommand{\tikzcuboid}[1]{
    \tikzset{cuboid,#1} 
  \pgfmathsetlengthmacro{\vectorxx}{\tikzcuboidkey{scalex}*cos(\tikzcuboidkey{anglex})*28.452756}
  \pgfmathsetlengthmacro{\vectorxy}{\tikzcuboidkey{scalex}*sin(\tikzcuboidkey{anglex})*28.452756}
  \pgfmathsetlengthmacro{\vectoryx}{\tikzcuboidkey{scaley}*cos(\tikzcuboidkey{angley})*28.452756}
  \pgfmathsetlengthmacro{\vectoryy}{\tikzcuboidkey{scaley}*sin(\tikzcuboidkey{angley})*28.452756}
  \pgfmathsetlengthmacro{\vectorzx}{\tikzcuboidkey{scalez}*cos(\tikzcuboidkey{anglez})*28.452756}
  \pgfmathsetlengthmacro{\vectorzy}{\tikzcuboidkey{scalez}*sin(\tikzcuboidkey{anglez})*28.452756}
  \begin{scope}[xshift=\tikzcuboidkey{shiftx}, yshift=\tikzcuboidkey{shifty}, scale=\tikzcuboidkey{scale}, rotate=\tikzcuboidkey{rotation}, x={(\vectorxx,\vectorxy)}, y={(\vectoryx,\vectoryy)}, z={(\vectorzx,\vectorzy)}]
    \pgfmathsetmacro{\steppingx}{1/\tikzcuboidkey{densityx}}
  \pgfmathsetmacro{\steppingy}{1/\tikzcuboidkey{densityy}}
  \pgfmathsetmacro{\steppingz}{1/\tikzcuboidkey{densityz}}
  \newcommand{\dimx}{\tikzcuboidkey{dimx}}
  \newcommand{\dimy}{\tikzcuboidkey{dimy}}
  \newcommand{\dimz}{\tikzcuboidkey{dimz}}
  
\pgfmathsetmacro{\xrangeend}{floor(\dimx/\steppingx)*\steppingx}
\pgfmathtruncatemacro{\xsteps}{\xrangeend/\steppingx}
\pgfmathsetmacro{\yrangeend}{floor(\dimy/\steppingy)*\steppingy}
\pgfmathtruncatemacro{\ysteps}{\yrangeend/\steppingy}
\pgfmathsetmacro{\zrangeend}{floor(\dimz/\steppingz)*\steppingz}
\pgfmathtruncatemacro{\zsteps}{\zrangeend/\steppingz}
\pgfmathsetmacro{\zcut}{\tikzcuboidkey{zcut}}

\foreach \xi in {1,...,\xsteps} {
  \foreach \yi in {1,...,\ysteps} {
    \pgfmathsetmacro{\x}{\xi*\steppingx}
    \pgfmathsetmacro{\y}{\yi*\steppingy}
    \pgfmathsetmacro{\lowx}{\x - \steppingx}
    \pgfmathsetmacro{\lowy}{\y - \steppingy}
    \filldraw[cuboid/front] (\lowx,\lowy,\dimz) -- (\lowx,\y,\dimz) -- (\x,\y,\dimz) -- (\x,\lowy,\dimz) -- cycle;
  }
}

\foreach \xi in {1,...,\xsteps} {
  \foreach \zi in {1,...,\zsteps} {
    \pgfmathsetmacro{\x}{\xi*\steppingx}
    \pgfmathsetmacro{\z}{\zi*\steppingz}
    \pgfmathsetmacro{\lowx}{\x - \steppingx}
    \pgfmathsetmacro{\lowz}{\z - \steppingz}
    \pgfmathparse{\z > \zcut}
    \ifnum\pgfmathresult=1
        \filldraw[cuboid/top] (\lowx,\dimy,\lowz) -- (\lowx,\dimy,\z) -- (\x,\dimy,\z) -- (\x,\dimy,\lowz) -- cycle;
    \else
        \filldraw[cuboid/topcut] (\lowx,\dimy,\lowz) -- (\lowx,\dimy,\z) -- (\x,\dimy,\z) -- (\x,\dimy,\lowz) -- cycle;
    \fi
  }
}

\foreach \yi in {1,...,\ysteps} {
  \foreach \zi in {1,...,\zsteps} {
    \pgfmathsetmacro{\y}{\yi*\steppingy}
    \pgfmathsetmacro{\z}{\zi*\steppingz}
    \pgfmathsetmacro{\lowy}{\y-\steppingy}
    \pgfmathsetmacro{\lowz}{\z-\steppingz}
    \pgfmathparse{\z > \zcut}
    \ifnum\pgfmathresult=1
        \filldraw[cuboid/right] (\dimx,\lowy,\lowz) -- (\dimx,\lowy,\z) -- (\dimx,\y,\z) -- (\dimx,\y,\lowz) -- cycle;
    \else
        \filldraw[cuboid/rightcut] (\dimx,\lowy,\lowz) -- (\dimx,\lowy,\z) -- (\dimx,\y,\z) -- (\dimx,\y,\lowz) -- cycle;
    \fi
  }
}

  \ifcuboidemphedge
    \draw[cuboid/emphstyle] (0,\dimy,0) -- (\dimx,\dimy,0) -- (\dimx,\dimy,\dimz) -- (0,\dimy,\dimz) -- cycle;%
    \draw[cuboid/emphstyle] (0,\dimy,\dimz) -- (0,0,\dimz) -- (\dimx,0,\dimz) -- (\dimx,\dimy,\dimz);%
    \draw[cuboid/emphstyle] (\dimx,\dimy,0) -- (\dimx,0,0) -- (\dimx,0,\dimz);%
  \fi

  \ifcuboidshade
    \pgfmathsetmacro{\cstepx}{\dimx/\tikzcuboidkey{shadesamples}}
    \pgfmathsetmacro{\cstepy}{\dimy/\tikzcuboidkey{shadesamples}}
    \pgfmathsetmacro{\cstepz}{\dimz/\tikzcuboidkey{shadesamples}}
    \foreach \s in {1,...,\tikzcuboidkey{shadesamples}}
    {   \pgfmathsetmacro{\lows}{\s-1}
        \pgfmathsetmacro{\cpercent}{(\lows)/(\tikzcuboidkey{shadesamples}-1)*100}
        \fill[opacity=\tikzcuboidkey{shadeopacity},color=\tikzcuboidkey{shadecolorlight}!\cpercent!\tikzcuboidkey{shadecolordark}] (0,\s*\cstepy,\dimz) -- (\s*\cstepx,\s*\cstepy,\dimz) -- (\s*\cstepx,0,\dimz) -- (\lows*\cstepx,0,\dimz) -- (\lows*\cstepx,\lows*\cstepy,\dimz) -- (0,\lows*\cstepy,\dimz) -- cycle;
        \fill[opacity=\tikzcuboidkey{shadeopacity},color=\tikzcuboidkey{shadecolorlight}!\cpercent!\tikzcuboidkey{shadecolordark}] (0,\dimy,\s*\cstepz) -- (\s*\cstepx,\dimy,\s*\cstepz) -- (\s*\cstepx,\dimy,0) -- (\lows*\cstepx,\dimy,0) -- (\lows*\cstepx,\dimy,\lows*\cstepz) -- (0,\dimy,\lows*\cstepz) -- cycle;
        \fill[opacity=\tikzcuboidkey{shadeopacity},color=\tikzcuboidkey{shadecolorlight}!\cpercent!\tikzcuboidkey{shadecolordark}] (\dimx,0,\s*\cstepz) -- (\dimx,\s*\cstepy,\s*\cstepz) -- (\dimx,\s*\cstepy,0) -- (\dimx,\lows*\cstepy,0) -- (\dimx,\lows*\cstepy,\lows*\cstepz) -- (\dimx,0,\lows*\cstepz) -- cycle;
    }
  \fi 
  \end{scope}
}

\title{The Universal Classifier for Graph Learning}

\author{%
  Ben Finkelshtein\\
  Google Research \\
  \And
  Andr\'{e} Linhares \\
  Google Research \\
  \And
  Petar Veli\v{c}kovi\'{c} \\
  Google DeepMind \\
  \AND
  \makebox[0pt]{}\vspace{-3em}
  \And
  Bryan Perozzi \\
  Google Research \\
  \And
  Mikhail Galkin \\
  Google Research \\
  \And
  \makebox[0pt]{}\vspace{-3em}
}

\iclrfinalcopy 
\begin{document}

\maketitle

\begin{abstract}
\looseness=-1
While foundation models have revolutionized natural language processing and computer vision by leveraging universal vocabularies, Graph Machine Learning (GML) remains fractured due to the absence of a unified feature and structural representation across diverse domains. 
Existing works claiming to be Graph Foundation Models (GFMs) are typically restricted to node-level predictions or require fixed feature dimensions, failing to provide a truly task-agnostic backbone for the full spectrum of graph learning applications.
In this paper, we introduce the \textbf{\protect\method (\protect\uc)}, which supports arbitrary feature and class cardinalities, \protect unifying node-, edge-, and graph-level objectives under a single similarity-based classification objective.
The \protect\uc reformulates all node-, edge-, and graph-level prediction tasks as maximizing similarity in the latent space: by lifting heterogeneous features and labels into 3D latent tensors, the model learns transferable features independent of specific input schemas.
This architecture allows a single pre-trained model to generalize to node classification, node regression, and link prediction across unseen graphs with varying feature semantics.
Experiments show strong zero-shot transfer performance across node-, link-, and graph-level tasks.
\end{abstract}

\section{Introduction}

\looseness=-1
The emergence of foundation models has fundamentally redefined the landscape of machine learning, demonstrating an unprecedented capacity for zero-shot generalization and cross-domain transfer in natural language processing \citep{raffel2023exploringlimitstransferlearning, touvron2023llamaopenefficientfoundation} and  computer vision \citep{radford2021learningtransferablevisualmodels, dosovitskiy2021imageworth16x16words}. However, the transition of this "foundational" paradigm to Graph Machine Learning (GML) has proven uniquely difficult due to two fundamental barriers.

First, unlike language or vision models that rely on a shared vocabulary of tokens or pixels, graphs exhibit extreme attribute heterogeneity; a node feature might represent a chemical property in a molecular graph or a textual embedding in a social network. This diversity necessitates schema invariant architectures capable of operating across arbitrary feature and label spaces \citep{zhao2025fullyinductivenodeclassificationarbitrary, finkelshtein2025equivariance}. Second, whereas traditional foundation models leverage a single
training objective like next-token prediction or masked patch reconstruction, the GML landscape spreads across a wide array of node-, edge-, and graph-level objectives that usually require bespoke architectures. 
While the GML community has progressed in building models that address these challenges independently, a true Graph Foundation Model must encapsulate both properties.

\begin{figure}[t]
    \centering
    \resizebox{0.92\textwidth}{!}{%
        \input{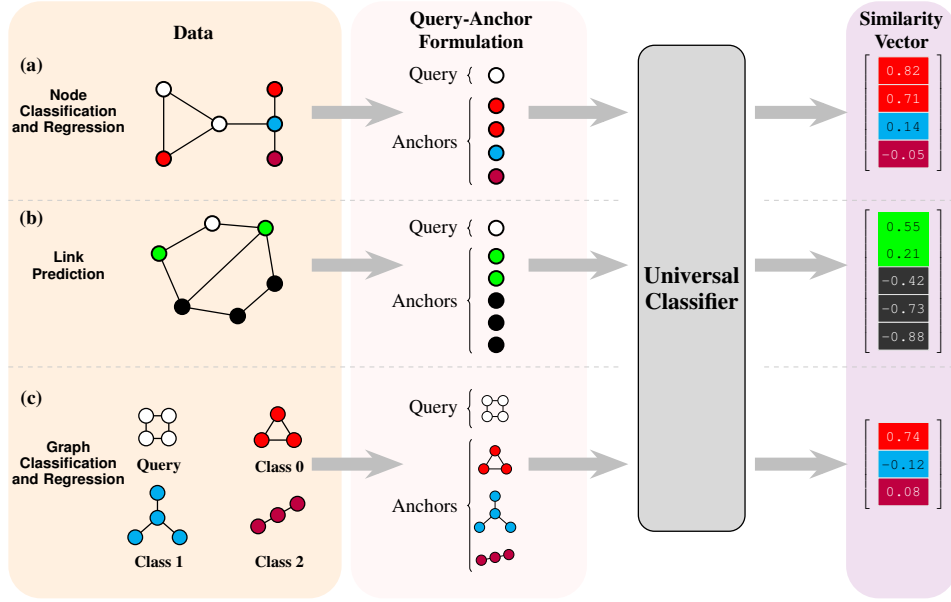} 
    }
    \captionsetup{justification=justified, singlelinecheck=false}
    \caption{Overview of the \method generalizing across tasks. \textbf{(a) Node Classification and Regression:} a query node is compared to a support set of anchors, with similarities aggregated by class. In regression, continuous targets are discretized into bins, treated as distinct classes, and recovered via similarity-weighted interpolation. \textbf{(b) Link Prediction:} a source node acts as a query and is evaluated against target nodes. Nodes connected to the query serve as anchors of a \emph{positive} class, while all other nodes act as anchors of a \emph{negative} class. \textbf{(c) Graph Classification and Regression:} a source graph is represented by pooling its respective features into a single global query, which is then compared against anchor graphs representing specific labels, similar to node-level tasks.}
    \label{fig:multi_task}

    \nointerlineskip
    \vbox to 0pt{\begin{subfigure}{0\textwidth}
        \phantomsubcaption\label{fig:node_class}
        \phantomsubcaption\label{fig:link_pred}
        \phantomsubcaption\label{fig:graph_class}
    \end{subfigure}\vss}%

\end{figure}

\begin{center}
    \definecolor{light-gray}{gray}{0.90}
        \tcbset{colback=light-gray,colframe=black,arc=0mm,boxrule=0.5pt}
        \begin{tcolorbox}
        \looseness=-1
        \textbf{Graph Foundation Model (GFM).} 
        A single pre-trained model capable of performing inference across arbitrary feature schemas while simultaneously addressing node-, edge-, and graph-level tasks across unseen graphs, without task-specific architectural modifications.
    \end{tcolorbox}
\end{center}

We introduce this terminology to bring clarity to a field where "GFM" is frequently used as a misnomer. Current literature offers a fractured landscape where models often claim the GFM title while satisfying only a subset of its essential properties. For instance, recent advancements in schema-invariant models \citep{zhao2024gcope,zhao2025fullyinductivenodeclassificationarbitrary, finkelshtein2025equivariance} 
are strictly confined to node classification. 
Similarly, generalist models for link prediction~\citep{lee2023ingraminductiveknowledgegraph,  galkin2024foundationmodelsknowledgegraph,huang2026hyper,kim2026flock} can hardly generalize to node- and graph-level tasks. 

In this paper, we unify discriminative prediction tasks on graphs under a single framework. To this end, we introduce the \method (\uc), a \emph{schema-invariant} model, which generalizes to arbitrary feature and label spaces and performs node-, edge-, and graph-level tasks without task-specific treatment.
\uc blends representation learning with in-context learning and reformulates all prediction tasks as similarity in a joint latent space. 
Specifically, \uc treats a set of labeled items (that could be nodes, edges, or graphs) as \emph{anchors}, and probes them against unlabeled items, \emph{queries}. 
By lifting anchors and queries to the shared latent space through expressive encoders (that mix features, labels, and graph topology),  \uc estimates how similar a query is to each of the anchors. 
This formulation allows \uc to unify all graph learning tasks to be trained with a single cross-entropy loss, lending it the name the \emph{universal classifier}.

\looseness=-1
\textbf{Unifying graph learning tasks.} \uc encodes each feature independently, representing each node's embedding as a lifted matrix in $\mathbb{R}^{F \times D}$. Depending on the specific task, \uc dynamically casts these representations as either \emph{query} or \emph{anchor} embeddings. This formulation unifies all tasks by identifying semantic relationships through pairwise similarities in a shared latent space. See  \cref{fig:multi_task} for details.

Our query-anchor paradigm naturally extends to multi-relational domains like knowledge graphs and relational databases, as well as to temporal and heterogeneous graphs, which we leave for future work. Our core contributions are:

\begin{itemize}[leftmargin=0.5cm, itemsep=0.5pt, topsep=0pt]
    \item \textbf{We frame all graph learning tasks as similarity estimation} via the schema-invariant encoder, applying cardinality independence jointly to features and labels so that a single parameter set generalizes to unseen feature schemas and arbitrary feature and class cardinalities.
    \item \looseness=-1 To our knowledge, \uc is the first model to perform node-, edge-, and graph-level tasks without architectural modifications, task-specific heads, or maximum feature and class dimensions.

    \item \uc demonstrates strong generalization performance in practical graph learning benchmarks. 
\end{itemize}

\section{The \method}
\label{sec:methodology}

\subsection{Notation}
\label{subsec:notations}

We denote $[i]=\{1, \ldots, i\}$ and the cardinality of a finite set $\gI$ by $|\gI|$. Vectors, matrices, and tensors are denoted in bold.  For any matrix or tensor $\mathbf{H} \in \sR^{N \times M\times K}$, we denote its $i$-th row (or slice) by $\mathbf{H}_{i}$. For a subset of indices $\gI$, the corresponding submatrix or subtensor is denoted as $\mathbf{H}_\gI \in \sR^{|\gI| \times M\times K}$.

\looseness=-1
We consider an undirected, unweighted graph $\mathcal{G} = (\mathcal{V}, \mathcal{E}, \mX, \mY)$, where $\mX \in \sR^{N \times F}$ is a matrix of node features, and $\mY$ denotes the labels. The structure of $\mY$ is task-dependent: for node classification, $\mY \in \{0,1\}^{N \times C}$ has one-hot rows $\mY_v$ over $C$ classes, while for node regression, $\mY \in \sR^{N\times 1}$ and each node label $\mY_v$ is a continuous scalar. The graph size is $|\mathcal{G}| = |\mathcal{V}|$. We denote the $k$-hop subgraph around a node $v$ by $\mathcal{G}^k(v)$. We also denote $N$ as the number of nodes, $F$ as the number of input features, $D$ as the hidden dimension, and $C$ as the number of output classes or regression targets.

\looseness=-1
The query-anchor pair $(\mathcal{Q}, \mathcal{A})$ unifies different graph learning objectives: $\mathcal{Q}$ represents a batch of query entities, and $\mathcal{A}$ represents a support set of labeled anchors. These entities are structurally flexible, capable of representing either individual nodes or entire graph topologies depending on the task.

\subsection{Data Sampling and Schema Invariance}
\label{subsec:data_handling}

\textbf{Sampling and Label Masking.} To prevent label leakage during training, we sample disjoint query ($\gQ$) and anchor ($\gA$) sets ($\gQ \cap \gA = \emptyset$). In the label matrix $\mY$, query labels are masked as zero-vectors, while unmasked labels are one-hot encoded. At inference, target nodes act as queries against the training set (or a subset), which serves as the anchor context. See Appendix~\ref{app:sampling} for more details.

\textbf{Feature Normalization.}  We first apply feature-wise z-normalization to raw node features $\mX$ to canonicalize the numerical range and provide a stable input distribution. Next, a shared MLP $\phi_{\text{feat}}: \sR \rightarrow \sR^{D}$ projects each normalized scalar feature independently, resulting in the 3D tensor:
\[
(\mX_{\text{init}})_{n,f,:} = \phi_{\text{feat}}(\mX_{n,f}), \quad \forall n \in [N],\, f \in [F], \qquad \text{so that } \qquad \mX_{\text{init}} \in \sR^{N \times F \times D}.
\]

\looseness=-1
Following \citep{finkelshtein2025equivariance,bevilacqua2025holographic}, treating the varying feature dimension $F$ as an unordered set enables schema invariance. 
All further \uc components operate along the fixed latent $D$ axis, decoupling model parameters from $F$ at both training and inference.

\looseness=-1
\textbf{Label Lifting for Node-level Tasks.}  Anchor labels can also serve as supplementary 3D node features. An additional learnable MLP $\phi_{\text{label}}: \sR \rightarrow \sR^{D}$ independently processes each entry of a node's label vector (whether a one-hot anchor or a zero-masked query). Similar to feature lifting, this maps the 2D label matrix $\mY \in \sR^{N \times C}$ to a 3D tensor $\mY_{\text{init}} \in \sR^{N \times C \times D}$ where, similarly to feature lifting, operating exclusively along the shared $D$ axis allows \uc to accommodate any number of classes $C$.

\subsection{Model Architecture}
\label{subsec:model_architecture}

\begin{figure}[t]
    \centering
    \resizebox{0.95\textwidth}{!}{%
        \input{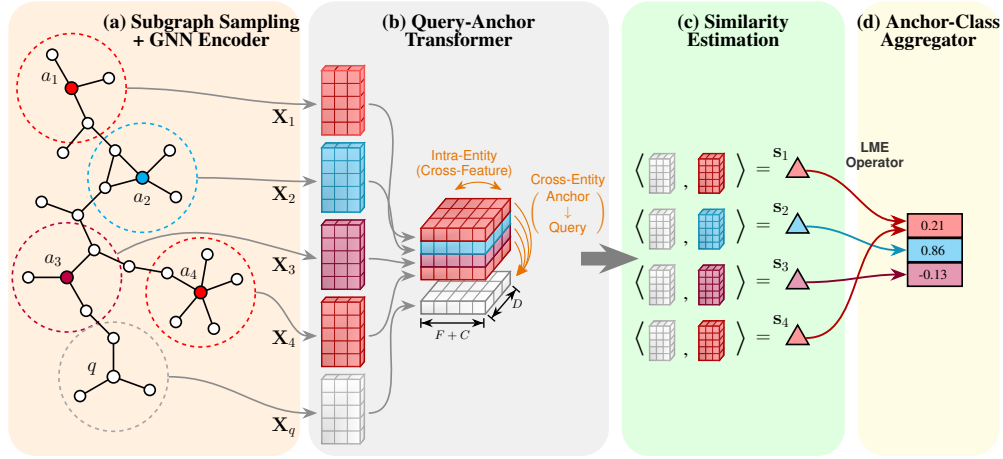} 
    }
    \captionsetup{justification=justified, singlelinecheck=false}
    \caption{The four primary modules of the \method. \textbf{(a) GNN Encoder} that processes the localized topology (obtained via subgraph sampling) of each query and anchor node. \textbf{(b) Query-Anchor Transformer} that performs intra-entity and cross-entity attention to contextualize the query within the anchor set. \textbf{(c) Similarity Estimator} that aggregates the representations of queries and anchors to produce query-anchor similarity scores. \textbf{(d) Anchor-Class Aggregator} that pools those pairwise similarity scores to produce probabilistic class predictions.}
    \label{fig:pipeline}
    \nointerlineskip
    \vbox to 0pt{\begin{subfigure}{0\textwidth}
        \phantomsubcaption\label{fig:subgraph_sampling_and_gnn_encoder}
         \phantomsubcaption\label{fig:query_anchor_transformer}
         \phantomsubcaption\label{fig:similarity_estimation}
         \phantomsubcaption\label{fig:anchor_class_aggregator}
    \end{subfigure} }%
    \vspace{-10pt}
\end{figure}

\looseness=-1
The \method includes four modules (\cref{fig:pipeline}): \textbf{(1) a GNN Encoder} that independently processes the localized topology of each query and anchor node; \textbf{(2) a Query-Anchor Transformer} that performs cross-feature and cross-anchor attention to contextualize the query within the anchor set; \textbf{(3) a Similarity Estimator} that aggregates these refined representations to produce the final query-anchor similarity scores; and \textbf{(4) an Anchor-Class Aggregator} that pools these pairwise scores by class to produce probabilistic class predictions.

\looseness=-1
\textbf{(1) Subgraph Sampling and GNN Encoder (\cref{fig:subgraph_sampling_and_gnn_encoder}).}
To ensure that query and anchor representations capture their local structural contexts, the forward pass begins by extracting an induced $k$-hop subgraph $\mathcal{G}^k(r)$ around each root node $r \in \gQ \cup \gA$, creating a unified graph $\mathcal{G}_{\text{uni}} = \bigcup_{r \in \gQ \cup \gA} \mathcal{G}^k(r)$. 
The shared $D$-dimensional GNN Encoder then performs message passing over the unified graph,
operating on the input 3D tensors (separately for feature $\mX$ and label $\mY$ representations):

\begin{align*}
    \mX^{l+1} = \texttt{GNN} \Big( \mX^{l}, \mathcal{G}_{\text{uni}}  \Big) \in \sR^{|\mathcal{G}_{\text{uni}}| \times F \times D} , \quad
    \mY^{l+1} = \texttt{GNN} \Big( \mY^{l}, \mathcal{G}_{\text{uni}}  \Big) \in \sR^{|\mathcal{G}_{\text{uni}}| \times C \times D}
\end{align*}

Following the final GNN layer,
we extract the representations of the root nodes 
$\{\mX_r, \mY_r\}_{r\in \gQ\cup\gA}$  
to serve as the final structure-induced feature and label representations for the queries and anchors.

\looseness=-1
\textbf{(2) Query-Anchor Transformer (\cref{fig:query_anchor_transformer}).} After extracting structural representations for each query and anchor, we seek to refine these embeddings by capturing their pairwise relational affinities. 
As we now operate on unordered sets of query and anchor representations, Transformers are natively suited for this task.
Self-attention without positional encodings is permutation-equivariant over a channel set whose size varies across datasets, whereas any mixer indexing channels by position would reintroduce the dependence on $F$ and $C$ that the lifting removes. We do not claim this design is optimal, but it is this property that makes it compatible with schema invariance.
Each $l$-th block performs two interleaved and orthogonal attention operations: \textit{Intra-Entity Attention} and \textit{Cross-Entity Attention}.

\looseness=-1
\textit{Intra-Entity Attention:} This mechanism operates across the $(F+C)$ channel axis to capture internal correlations between the features and labels of the \emph{same node}. Concatenating the two matrices allows a query's value on one attribute to be read together with the labels of the anchors sharing that pattern; carried separately, no relationship between a feature and a label could be represented. For a specific query or anchor $r\in \gQ\cup\gA$, self-attention is applied directly to the concatenated matrices in $\sR^{(F+C)\times D}$:

\begin{equation*}
    [\Bar\mX_r^l, \Bar\mY_r^l] = \texttt{TransformerBlock}\left([\mX_r^l \parallel \mY_r^l], \: [\mX_r^l \parallel \mY_r^l]\right) \in \sR^{(F+C) \times D}
\end{equation*}

\looseness=-1
where $\parallel$ denotes concatenation along the feature-label axis, and the \emph{query-key} notation highlights the self-attention nature of the function.  
We split $\Bar\mX_r^l$ into query representations $\Bar\mX_{\gQ}^l \in \sR^{|\gQ| \times F \times D}$ and $\Bar\mY_{\gQ}^l \in \sR^{|\gQ| \times C \times D}$ and anchor representations $\Bar\mX_{\gA}^l \in \sR^{|\gA| \times F \times D}$ and $\Bar\mY_{\gA}^l \in \sR^{|\gA| \times C \times D}$, respectively.

\textit{Cross-Entity Attention:} This mechanism operates across the query-anchor dimension, allowing queries to gather relevant context by directly comparing their own features and labels to the anchor set, that is, the query set $\gQ$ attends directly to the corresponding channel across the anchor set $\gA$:

\begin{align*}
    \mX_{\gQ}^{l+1} = \texttt{TransformerBlock}\left(\Bar\mX_{\gQ}^l, \Bar\mX_{\gA}^l\right), \quad
    \mY_{\gQ}^{l+1} = \texttt{TransformerBlock}\left(\Bar\mY_{\gQ}^l, \Bar\mY_{\gA}^l\right) 
\end{align*}

\looseness=-1
Here, queries are query representations, while keys and values are anchor representations. 
Only queries are updated by this step; the anchors carry their intra-entity representations, as is standard in tabular foundation models \citep{tabpfnv2,tabicl2025}. See  Appendix~\ref{app:sampling} for details.

\looseness=-1
\textbf{(3) Similarity Estimation (\cref{fig:similarity_estimation}).} Having mapped the features and labels of each query and anchor into a shared latent space, we quantify the relational affinity between each query $q \in \gQ$ and anchor $a \in \gA$ by computing a scalar similarity score $s(q, a) \in \sR$ using both feature and label representations.
The scoring function can be parameterized by another neural net or can be non-learnable, such as a scaled dot product. 
We employ a scaled Frobenius inner product similarity, calculated as

\begin{equation*}
    s(q, a) = \frac{1}{2D} \left( \frac{1}{F} \langle \mX_{q}, \mX_{a} \rangle_{\text{Frob}} + \frac{1}{C} \langle \mY_{q}, \mY_{a} \rangle_{\text{Frob}} \right)
\end{equation*}

\looseness=-1
where $\langle \cdot, \cdot \rangle_{\text{Frob}}$ denotes the Frobenius inner product (a dot product over flattened matrices). By normalizing over the hidden dimensionality $D$ and averaging over the channel sets, we ensure the resulting similarity metric remains numerically stable and invariant to the input data cardinality ($F$ and $C$).

\looseness=-1
\textbf{(4) Anchor-Class Aggregator (\cref{fig:anchor_class_aggregator}).} To derive class-level predictions, we aggregate the pairwise similarity scores across anchors by their labels. The aggregation should be robust to class imbalance and produce logits for a standard cross-entropy loss. An operation that achieves both is the \textbf{LogMeanExp} (LME) operator. Given a single query with a vector of similarity scores $\vs \in \sR^{|\mathcal{A}|}$, where each score corresponds to the similarity with one of the anchors, and a binary mask matrix $\mM \in \{0, 1\}^{C\times |\mathcal{A}|}$ indicating the class each anchor is affiliated with, the class-level logit $\vz_c$ is defined 
as:

\[
  \vz_c = \operatorname{LME}(\vs, \mM)_c =
  \begin{cases}
    \displaystyle \log \left( \frac{1}{\|\mM_c\|_1} \mM_c \exp(\vs) \right), & \text{if } \|\mM_c\|_1 > 0, \\[8pt]
    -\infty, & \text{otherwise,}
  \end{cases}
  \quad \forall c \in [C]
\]

\looseness=-1
where $\|\mM_c\|_1 = \sum_{a=1}^{|\mathcal{A}|} |\mM_{c,a}|$ denotes the total number of anchors belonging to class $c$. By explicitly dividing by the class support size, this operator ensures that the contextual contribution of any given class is determined by its average anchor likelihood, preventing imbalanced classes from dominating the summation. The predicted class is then the index of the maximum logit, $y_{\text{pred}} = \arg\max_c \vz_c$.

\subsection{Node Regression}
\label{subsec:regression}

\looseness=-1
We follow a common protocol to convert regression into classification~\citep{stewart2025building} via two deterministic transformations of the target space. \textbf{Input Discretization} first transforms continuous values into soft bins, while \textbf{Output Interpolation} maps the model's aggregated classification logits from bin centroids back into scalars. Neither introduces learned parameters or a separate regression head.

\textbf{Input Discretization.} Given the continuous target matrix $\mY \in \sR^{N \times 1}$, we partition its value range into $C$ discrete bins\footnote{The number of bins and binning strategies are not fixed and can vary for each dataset.} defined by their edges $\ve = (e_0, \dots, e_C) \in \sR^{C+1}$. 
Rather than assigning each node $n \in [N]$ to a single hard bin, we transform its scalar target $\mY_{n,0} \in \sR$ into a soft probability distribution $\mY'_{n,:} \in [0, 1]^C$. Since a hard assignment would treat the bins as unordered, independent categories and erase the numerical distance between them, our soft approach explicitly preserves the ordinal relationships. This ensures the model understands that adjacent bins are numerically closer to one another than distant bins. This is achieved via an exponential softening function:
\[
    \mY'_{n,c} = \frac{\exp(-\tau | \mu_c - \mY_{n,0} |)}{\sum_{j=1}^C \exp(-\tau | \mu_j - \mY_{n,0} |)}, \quad \forall n \in [N], c \in [C]
\]

\looseness=-1
where $\mu_c = \frac{e_{c-1} + e_c}{2}$ represents the centroid of bin $c$, and $\tau$ is a temperature controlling the concentration of the distribution. These soft labels are then processed as in classification, lifting each scalar into the initial 3D label tensor via the shared MLP $(\mY_{\text{init}})_{n,c,:} = \phi_{\text{label}}(\mY'_{n,c})$ for all $n \in [N], c \in [C]$ --- the same $\phi_{\text{label}}$ as in \cref{subsec:data_handling}, whose parameters are independent of the bin count $C$.

\textbf{Output Interpolation.}
The main classification pipeline
outputs a vector of class-level logits $\vz \in \sR^C$ for a given query. At the output stage, 
we recover the final continuous prediction $y_{\text{pred}}$ by computing a similarity-weighted average over all bin centroids
$ y_{\text{pred}} = \sum_{c=1}^C \operatorname{softmax}(\vz)_c \mu_c. $
By allowing the query prediction to be interpolated across a range of relevant bin centroids, this expectation accurately captures the underlying continuous distribution. 
Furthermore, this architectural unification allows the model to be optimized using Kullback-Leibler (KL) divergence between the soft target distributions $\mY'$
generated during the discretization phase and the predicted output probabilities $\operatorname{softmax}(\vz)$.
This seamlessly aligns with our  classification framework:
minimizing KL divergence is fundamentally equivalent (the gradients are identical) to minimizing cross-entropy for soft labels, and strictly reduces to the standard cross-entropy objective when targets are one-hot encoded.

\subsection{Edge- and Graph-Level Tasks}
\label{subsec:link_graph_level}

\looseness=-1
\textbf{Link Prediction.} In this setting, the objective is to score edges rather than assign discrete categories. 
We adapt the core pipeline for this task with several modifications: \emph{(1)} we bypass the 3D label branch since there are no node-level classes to embed; \emph{(2)} we treat all nodes as candidate anchors, allowing the Similarity Estimator to directly output a pairwise similarity vector $\vs\in\sR^N$ for a given query, representing edge likelihoods (see \cref{fig:link_pred}); \emph{(3)} we remove the target edges from  message passing during both training and inference to prevent the GNN encoder from trivially memorizing the training topology; and \emph{(4)} we frame training as a binary classification task over the edge scores.

\looseness=-1
Given a batch of observed edges, we designate the source nodes as queries $\mathcal{Q}$,  the target nodes as positive anchors $\mathcal{A}_{\text{pos}}$, and augment this support set with negative anchors $\mathcal{A}_{\text{neg}}$ where $256$ negative anchors are sampled uniformly without replacement from non-neighbors in the training graph per positive edge. We then use LME to independently aggregate the similarities of the positive and negative anchor sets for each query. Finally, we apply cross-entropy over the two class logits.

\looseness=-1
\textbf{Graph-Level Tasks.} For graph classification and regression, the unit of representation is not a root node but the entire graph (see \Cref{fig:graph_class}). 
The core pipeline is adapted through: \emph{(1)} computing a single graph representation after the GNN encoder by mean pooling; \emph{(2)} injecting the 3D label tensor into the pooled graph-level representations before the Query-Anchor Transformer. Once these global embeddings are contextualized against the anchor graphs, similarity estimation and prediction proceed as in the node-level pipeline, with graph regression supported as in \Cref{subsec:regression}.

\section{Experiments}
\label{sec:experiments}

\looseness=-1
\textbf{Datasets.} 
Following recent works on the quality of  graph learning benchmarks~\citep{platonov2023critical,bechlerspeicher2025positiongraphlearninglose}, we evaluate the \uc across a diverse and representative corpus of real-world datasets spanning various input feature distributions, graph properties (e.g., sparse, dense, homophilic, heterophilic), and labels.
We partition our suite across four tasks: \textbf{(1) Node Classification}: \textit{arxiv}~\citep{hu2020open}, \textit{roman-empire}, \textit{amazon-ratings} \citep{platonov2023critical}, \textit{full-dblp}, \textit{full-cora} \citep{bojchevski2018deepgaussianembeddinggraphs}, \textit{hm-categories}, \textit{tolokers-2}, \textit{city-reviews}, and \textit{artnet-exp} \citep{bazhenov2024graphland}; \textbf{(2) Node Regression}: Random Low (RL) and High (RH) splits of \textit{city-roads-M}, \textit{city-roads-L}, \textit{twitch-views}, and \textit{artnet-views} \citep{bazhenov2024graphland}; \textbf{(3) Link Prediction}: \textit{roman-empire} \citep{platonov2023critical}, \textit{wiki-cs} \citep{mernyei2022wikicswikipediabasedbenchmarkgraph}, \textit{city-roads-M} and \textit{tolokers-2} \citep{bazhenov2024graphland}; and \textbf{(4) Graph Regression}: electronic-circuit datasets \textit{EC5}, \textit{EC7} and \textit{EC10} with efficiency (EFF) and output-voltage (VOUT) targets \citep{graphbench}.
We make all directed graphs undirected by adding inverse edges. Dataset statistics are presented in Appendix~\ref{app:datasets}.

\looseness=-1
\textbf{Pre-training.}
We train \uc on a mixture of openly available and synthetic datasets comprising node- and edge-level tasks.
While organizational policy prevents us from releasing the mixture, our training code, or the parameter count, we state the property on which the validity of our results depends: no evaluation dataset appears in the pre-training mixture --- no evaluation graph, no variant of one, and no split of one is seen under any task, either as a graph, in its features, or in its labels.

\looseness=-1
We evaluate a single model's out-of-the-box generalization via zero-shot (ZS) transfer, while assessing headroom performance via per-dataset fine-tuning (FT). Experiments were run on TPUv7.
Re-training a model several times solely to estimate initialization variance is prohibitively expensive, so we evaluate a single fixed checkpoint and report \uc results without error bars. Fine-tuning seed randomness stays below $0.1\%$ in most cases.

\subsection{Node Classification}

\textbf{Baselines.} We compare our framework against state-of-the-art inductive node classifiers 
GraphAny \citep{zhao2025fullyinductivenodeclassificationarbitrary} and TS-Mean \citep{finkelshtein2025equivariance}, along with recent methods inspired by tabular foundation models: NodePFN \citep{choi2026learning} and GraphPFN \citep{eremeev2026graphpfnpriordatafittedgraph}. We also compare against best-reported supervised GNN models trained end-to-end on each dataset in the full-batch mode to measure a possible performance headroom.
A fair comparison lies with methods that can be applied zero-shot to an unseen graph of arbitrary feature and label cardinality without fitting a dataset-specific output layer. 
Baseline results are sourced from the respective papers.

\begin{figure}[p]
    \centering
    \begin{subfigure}[b]{\textwidth}
        \textbf{(a) Node Classification}\par\smallskip
        \centering
        \includegraphics[width=\textwidth]{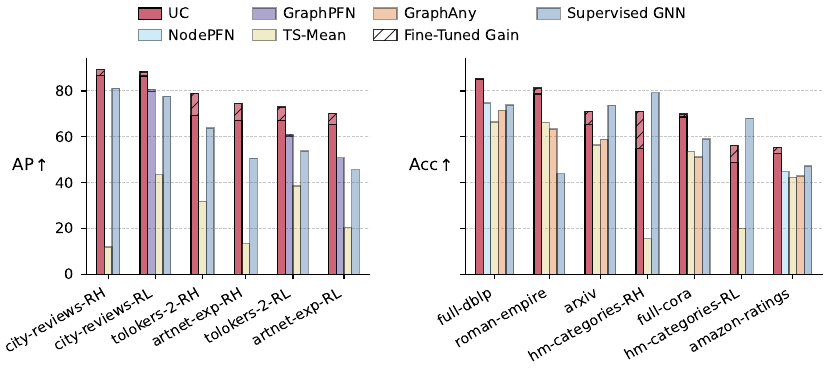}
    \phantomcaption\label{fig:nc}
    \end{subfigure}

    \vspace{-1.6em}
    \begin{tikzpicture}
        \draw[dashed, black!20] (0,0) -- (\textwidth,0);
    \end{tikzpicture}
    \vspace{0.2em}

    \begin{subfigure}[b]{\textwidth}
        \textbf{(b) Node and Graph Regression}\par\smallskip
        \centering
        \includegraphics[width=\textwidth]{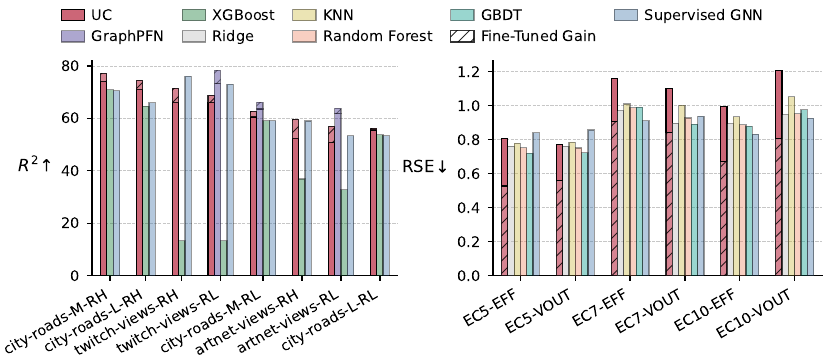}
    \phantomcaption\label{fig:regression}
    \end{subfigure}
    
    \vspace{-1.2em}
    \begin{tikzpicture}
        \draw[dashed, black!20] (0,0) -- (\textwidth,0);
    \end{tikzpicture}
    \vspace{0.1em}

    \begin{subfigure}[b]{\textwidth}
        \textbf{(c) Link Prediction}\par\smallskip
        \centering
        \includegraphics[width=0.5\textwidth]{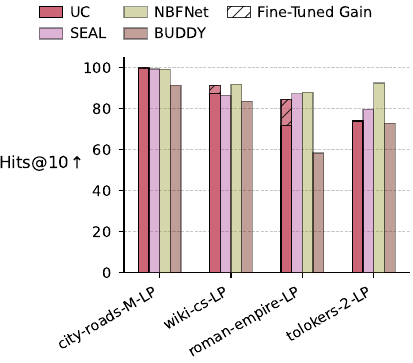}
    \phantomcaption\label{fig:link_prediction}
    \end{subfigure}

    \caption{The \uc performance across tasks: \textbf{(a)} binary (left, AP\,$\uparrow$) and multi-class (right, Accuracy\,$\uparrow$) node classification; \textbf{(b)} node (left, $R^2$\,$\uparrow$) and graph (right, RSE\,$\downarrow$) regression; and \textbf{(c)} link prediction (Hits@10\,$\uparrow$), comparing zero-shot transfer and fine-tuning to tabular FMs and supervised baselines.}
    \label{fig:main_results}
\end{figure}
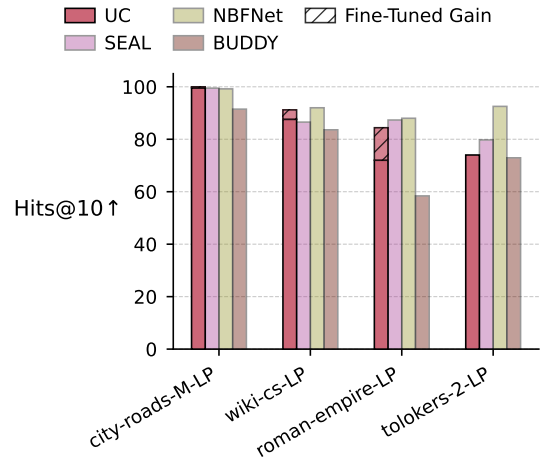

\looseness=-1
\textbf{Results and Observations.} 
The results in \Cref{fig:nc} demonstrate the \uc consistently outperforms both inductive classifiers and tabular FM-derived baselines on graphs with different structural, feature, and label properties. 
Further short dataset-specific fine-tuning enables even higher classification performance. 
Notably, in most cases, the \uc even surpasses supervised GNNs trained separately on each graph end-to-end. 
The single outlier on \emph{hm-categories} can be attributed to the dataset's larger average degree ($230+$) to which full-batch trained GNNs are more amenable. 

\subsection{Generalizing To Node Regression}

\looseness=-1
\textbf{Baselines.} We evaluate \uc against end-to-end supervised GNN and XGBoost \citep{Chen_2016} baselines. While included to establish a performance ceiling, comparing our zero-shot framework to models optimized on the target datasets is inherently disadvantaged. A fair comparison is GraphPFN \citep{eremeev2026graphpfnpriordatafittedgraph}, currently the only other zero-shot regression method.

\looseness=-1
\textbf{Results and Observations.} As shown in \cref{fig:regression}, framing regression as soft-binned classification via KL divergence is highly effective. The \uc remains highly competitive; matching or exceeding the transfer performance of specialized tabular foundation models. Notably, a key advantage of tabular-inspired baselines stems from their extensive pre-training on large mixtures of synthetic datasets. 
Upon fine-tuning, the \uc improves markedly across all datasets, narrowing the gap.

\subsection{Generalizing to Graph-Level Tasks}
\label{subsec:graph_level}

\looseness=-1
\textbf{Baselines.} To our knowledge, no generalist model performs graph regression on unseen graphs, so we compare the \uc against models trained per dataset: a supervised GNN and four classical regressors, Ridge, $k$-nearest neighbors (KNN), Random Forest, and gradient-boosted trees (GBDT). We report the relative squared error (RSE), where $1.0$ equals predicting the training mean.

\looseness=-1
\textbf{Results and Observations.} \Cref{fig:regression} shows that the fine-tuned \uc attains the lowest error on all six targets, without adding a graph-level head or any parameters. Remarkably, although the checkpoint was never pre-trained on a graph-level objective, the zero-shot \uc already matches the supervised GNN on EC-5.
We found discrepancies in the original EC dataset splits from ~\citet{graphbench} and re-ran the baselines, more details in Appendix~\ref{app:datasets}.

\subsection{Link Prediction as Classification}

\textbf{Baselines.} As no existing multi-task model performs link prediction without end-to-end training, we evaluate the \uc against end-to-end supervised link prediction models SEAL \citep{zhang2018linkpredictionbasedgraph}, NBFNet \citep{nbfnet}, and BUDDY \citep{chamberlain2023graphneuralnetworkslink} which establish a performance ceiling; comparing against them is inherently disadvantageous, but it contextualizes the out-of-the-box generalization capabilities of the \uc on unseen graphs.

\looseness=-1
\textbf{Evaluation Protocol.} We discard the labels and construct a link prediction task on each graph. Validation/test hold out $200/250$ edges with distinct and disjoint source nodes. Message passing uses training edges only at every stage, so no held-out edge is visible. Each positive is ranked against $200$ uniformly drawn candidate targets, with the source's true neighbors masked and ties broken pessimistically. All baselines are trained on the same subset of $10{,}000$ edges (further details in Appendix~\ref{app:datasets}).

\textbf{Results and Observations.} \Cref{fig:link_prediction} shows that the \uc generalizes to link prediction on unseen graphs: zero-shot, it outperforms BUDDY on every dataset and all baselines on \emph{city-roads-M}. Fine-tuning further narrows the gap, attaining the best result on \emph{city-roads-M} and approaching NBFNet on \emph{wiki-cs}; only on the dense \emph{tolokers-2} do path-based models retain an edge.

\subsection{Ablations and Qualitative Analysis}

\looseness=-1
\textbf{Ablations.} 
We probe the importance of certain components of the \uc architecture and re-train the entire model from scratch with ablated modules and report performance on \emph{roman-empire} (classification) and \emph{city-roads-M-RH} (regression). \Cref{tab:ablations} summarizes the results indicating that a shallow 2-layer Query-Anchor Transformer is detrimental, and 3-hop subgraph sampling offers no benefit over 2 hops. Removing or replacing whole modules is even more revealing, as no such variant approaches the full model. The GNN encoder contributes most: without it, performance on \emph{city-roads-M-RH} falls below the XGBoost and supervised GNN baselines (\cref{tab:detailed_nr}), indicating that the structure does substantial work rather than decorating a tabular pipeline. Replacing the LME aggregator with simple averaging and removing the tabular encoder both degrade performance, the former slightly more on both tasks.

\begin{wraptable}[11]{r}{0.4\columnwidth}
    \centering
    \caption{\uc architectural ablations.}
    \label{tab:ablations}
    \resizebox{\linewidth}{!}{
        \begin{tabular}{l c c}
    \toprule
    Parameter & \makecell{\emph{roman-} \\ \emph{empire} (Acc)} 
    & \makecell{\emph{city-roads-} \\ \emph{M-RH} ($R^2$)} \\
    \midrule
    Base \uc  & 78.61 & 74.08\\
    - 2 Trf layers  & 75.63 & 70.90 \\
    - 3 hops  & 77.25 & 67.94\\
    - GNN encoder & 64.35 & 53.30\\
    - Tabular encoder & 74.27 & 73.11\\
    - Mean score agg & 72.83 & 71.79\\
    \bottomrule
\end{tabular}
    }
\end{wraptable}

\begin{figure}[t]
    \centering
    \begin{subfigure}[b]{0.32\textwidth}
        \centering
        \includegraphics[width=\textwidth]{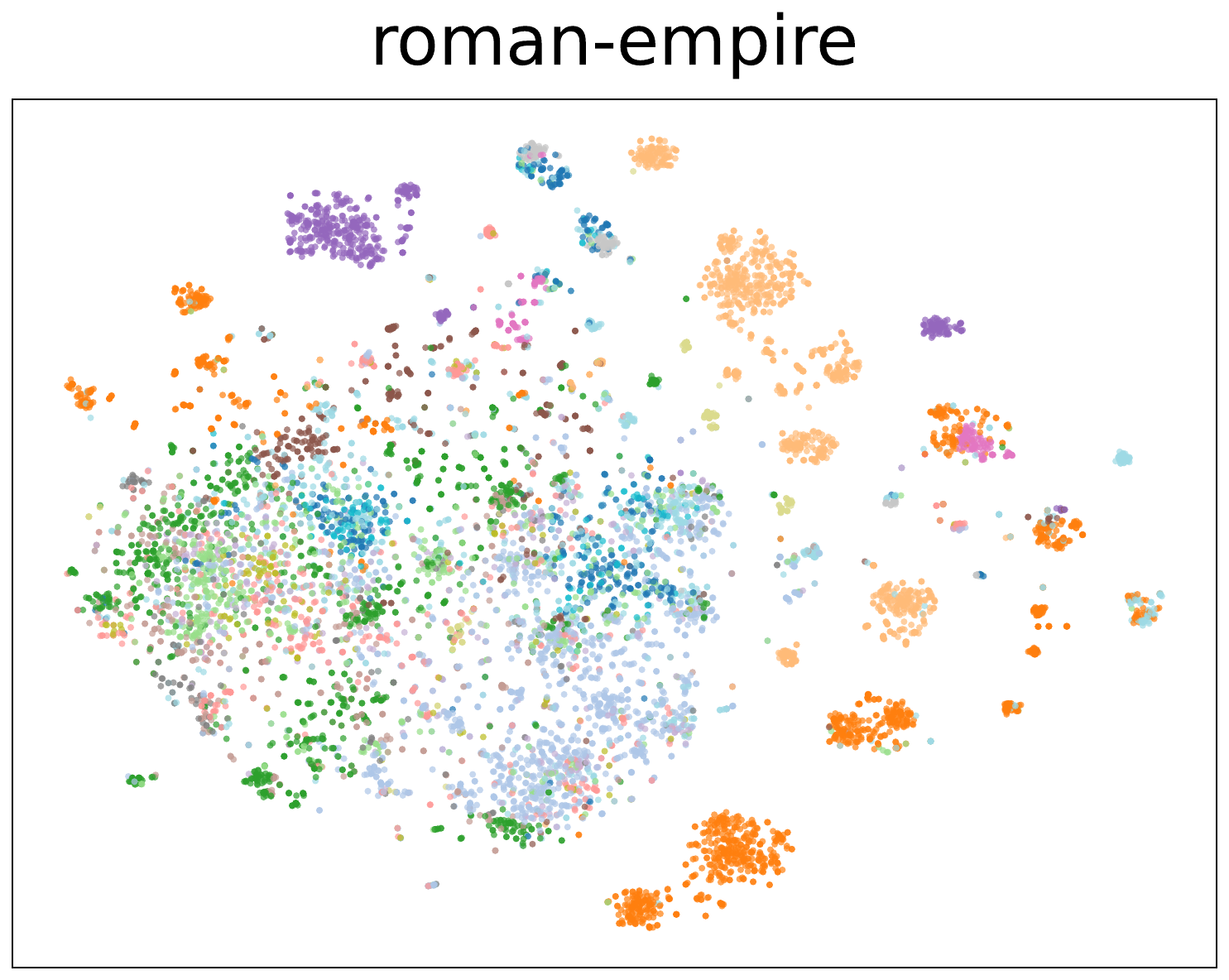}
    \end{subfigure}
    \hfill
    \begin{subfigure}[b]{0.32\textwidth}
        \centering
        \includegraphics[width=\textwidth]{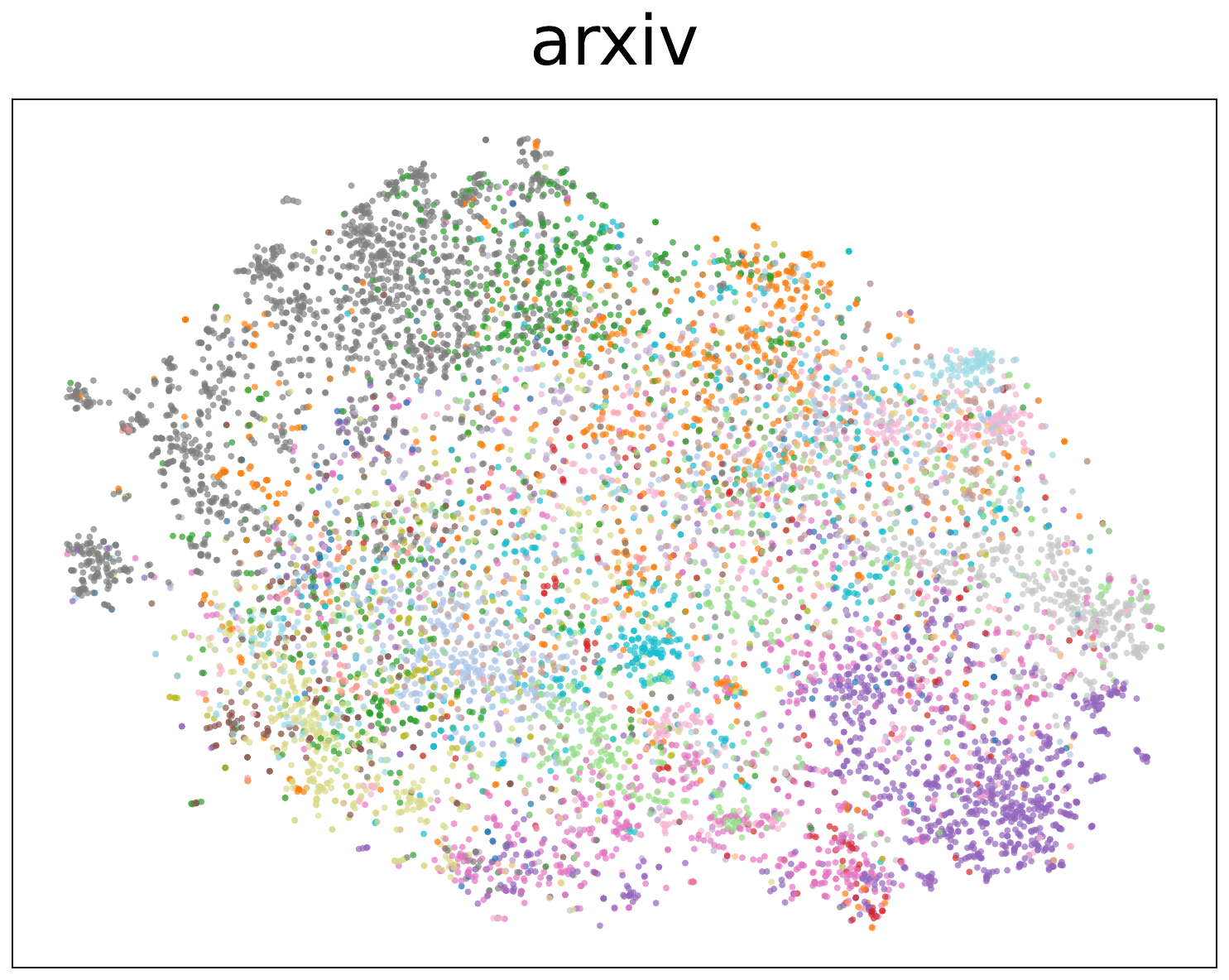}
    \end{subfigure}
    \hfill
    \begin{subfigure}[b]{0.32\textwidth}
        \centering
        \includegraphics[width=\textwidth]{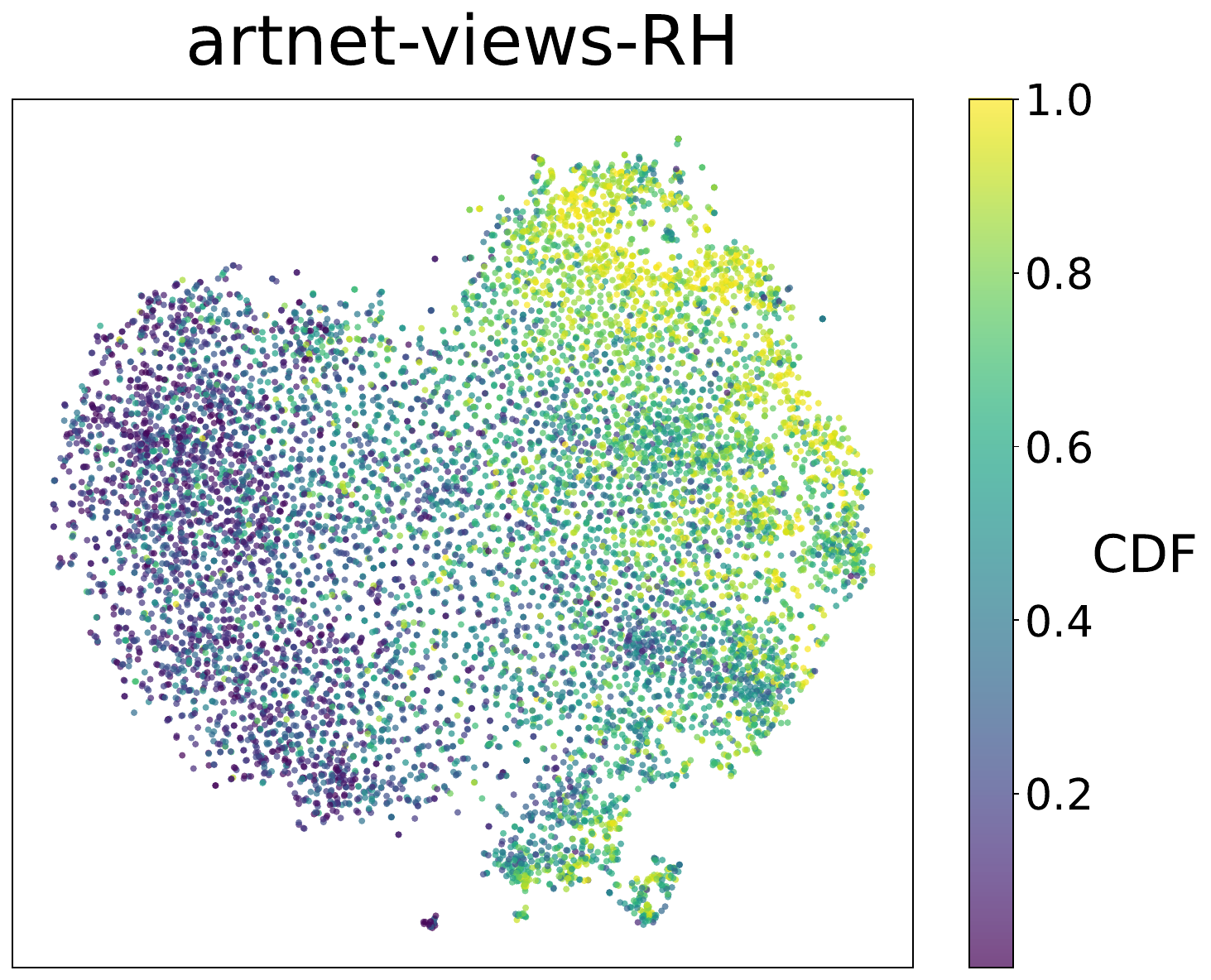}
    \end{subfigure}

    \caption{t-SNE projections of \uc anchor representations for node classification (\emph{roman-empire, arxiv}) and regression (\emph{artnet-views-RH}) datasets cluster around the labels. In regression datasets, labels are ordinal bins constructed around centroids, and nearby bins cluster together.}
    \label{fig:tsne}
\end{figure}

\looseness=-1
\textbf{Peeking into representations.} The \uc is a representation learning method and we examine the quality of anchor representations (coming from the training set) $\mX_{\mathcal{A}}$ of different classification and regression datasets. 
When passed through the GNN and Query-Anchor Transformer encoders and projected with t-SNE \citep{van2008tsne}, we observe (\Cref{fig:tsne}) the representations to form clusters indicative of the node labels both for heterophilic \emph{roman-empire} and homophilic \emph{arxiv}. 
When applied to regression (\emph{artnet-views-RH)}, continuous labels are discretized into bins that have an inherent ordinal structure, e.g., a centroid of $i$-th bin $\mu_i$ is numerically closer to $\mu_{i-1}$ than to $\mu_{i-5}$. 
\Cref{fig:tsne} demonstrates that the representations successfully capture this ordinality: bins corresponding to lower target values (the bottom 40\%) and higher target values (the top 40\%) naturally organize into distinct clusters.

\begin{wrapfigure}[18]{r}{0.5\columnwidth}
    \centering
    \includegraphics[width=\linewidth]{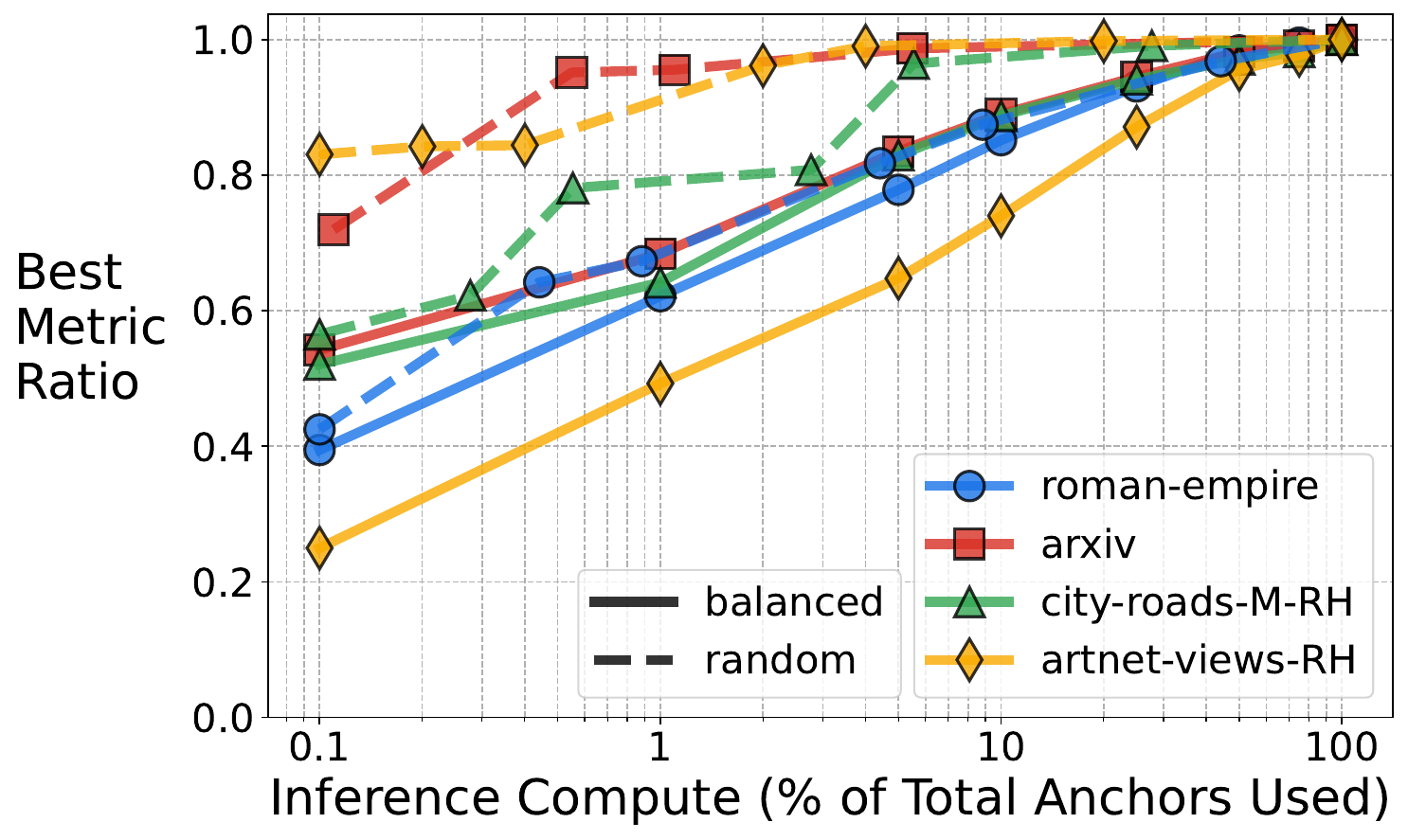}
    \caption{Inference-time performance (accuracy for classification and $R^2$ for regression) scaling with larger context sizes. 5\% of anchors is often enough to recover 80+\% of the final performance.}
    \label{fig:anchor_scaling}
\end{wrapfigure}

\looseness=-1
\textbf{Inference-time scaling}.
The \uc is also an in-context learning method and we posit that more context anchors improve performance at the cost of inference-time compute. 
To measure the impact of the context size, we probe downstream classification and regression performance by progressively reducing the support set of anchors. 
We employ two strategies for anchor sampling: \emph{balanced} with the same number of anchors per class; and \emph{random} where class sizes may differ.
Results indicate (\Cref{fig:anchor_scaling}) that the \uc benefits from inference-time compute on both classification and regression but is still able to recover $80+\%$ of the final performance in extremely compute-limited cases where only $1-5\%$ of anchors are present, thus saving $20-100\times$ flops of compute. 
Notably, the \emph{random} anchor sampling strategy is overall more effective and, e.g., recovers 90+\% of the final performance with $<1\%$ of anchors on \emph{arxiv} and \emph{artnet-views-RH}.
\section{Related Work}
\label{sec:related}

\textbf{Inductive Node Classifiers.}
A handful of GNNs~ \citep{zhao2024gcope,zhao2025fullyinductivenodeclassificationarbitrary,finkelshtein2025equivariance} have been proposed for performing node classification on unseen graphs. 
Typically, such methods rely on classification-specific feature and class preprocessing, do not leverage representation learning, do not show benefits of scaling data and compute, and cannot generalize to node regression nor link-, and graph-level tasks. 
While
some recent models~\citep{bechlerspeicher2026billionscalegraphfoundationmodels} extend to both node classification and link prediction, they remain bound to one graph and fixed input feature dimensions.

\textbf{Knowledge Graph FMs.}
KGs are graphs coming (in most cases) without node features and only having distinct edge types. 
The meaningful tasks on such graphs revolve around single- and multi-hop KG completion (link prediction) and such a constrained setup was amenable for the first attempts at creating generalist, inductive link predictors~\citep{lee2023ingraminductiveknowledgegraph,galkin2024foundationmodelsknowledgegraph, galkin2024ultraquery, zhang2024trix, cui2024kgicl,  huang2025expressiveknowledgegraphfoundation,kim2026flock}, which kickstarted the community interest in GFMs.
Despite the fast progress and extensions to novel structures like relational hypergraphs~\citep{huang2026hyper}, such models can hardly generalize to node- and graph-level tasks, cannot deal with arbitrary features, and do not seem to benefit from scaling (as leveraging pure structural patterns is often enough for high link prediction performance).

\looseness=-1
\textbf{Tabular and Relational FMs.}
The success of tabular FMs~\citep{tabpfnv1,tabpfnv2,tabicl2025} and their temporal extensions~\citep{das2023timesfm, hoo2025tablestimetabpfnv2outperforms} inspired graph variants that encode the topology as additional Laplacian- or random walk-derived feature columns with non-parametric message passing~\citep{eremeev2025turningtabularfoundationmodels,hayler2025bringinggraphstablezeroshot}, sometimes trained on synthetic graphs~\citep{eremeev2026graphpfnpriordatafittedgraph,choi2026learning}. 
Beyond single tables, Relational Deep Learning processes multi-table databases as temporal, heterogeneous graphs linked by primary-foreign keys~\citep{dbinfer,dwivedi2025relationaldeeplearningchallenges}, and generalist models~\citep{hudovernik2026kumorfm2scalingfoundationmodels,xu2026no} either rely on the graph structure induced by the schema or adapt single-table FMs.
Albeit effective, such methods are confined to node classification and regression tasks and are generally agnostic to the input graph structure and, therefore, cannot perform link- or graph-level tasks out of the box.
We posit that tabular FMs are a strict subset of GFMs as they treat row-wise data as a set of disconnected nodes.

\looseness=-1
\textbf{Scalar Lifting and Label Handling.} Lifting a scalar into a shared latent axis is not specific to graphs: tabular foundation models (TFMs) encode individual table cells similarly \citep{tabpfnv2,tabicl2025}, and schema-invariant graph models \citep{bevilacqua2025holographic,finkelshtein2025equivariance} adopt the construction on the feature axis. The \uc departs from them in two respects.
First, it lifts labels alongside features, a trait shared only by \citet{finkelshtein2025equivariance}. 
Second, they differ in what follows the lifting. Prior graph constructions and TFMs let the label space set the output dimension, either through a prediction head whose width is the class count or by reading predictions directly off the lifted label tensor. 
The \uc does neither: an anchor's label is input information only, not a channel to project onto, and a prediction is a comparison, with each query scored against every anchor and those scores pooled over the anchors sharing a class (see \cref{subsec:model_architecture}).

\section{Conclusion and Future Work}
\label{sec:conclusion}

\looseness=-1
We introduced the Universal Classifier (\uc) that frames common graph learning tasks under the same classification objective. 
The \uc blends representation learning with the in-context learning mechanism built around \emph{queries} and \emph{anchors} that allows a single model to effectively generalize to node-, link-, and graph-level tasks.
Extensive experiments demonstrate that a single pre-trained \uc model can achieve strong zero-shot generalization across diverse, unseen graphs with different structural properties, feature and label distributions, and frequently outperforms dataset-optimized, task-specific architectures. 
More discussion on limitations and future work is in Appendix~\ref{app:future}.

\bibliography{iclr2027_conference}
\bibliographystyle{iclr2027_conference}

\appendix
\section{Time Complexity and Runtime}
\label{app:complexity}
We derive the asymptotic cost of a forward pass of the \uc, compare it to a standard message-passing network, and report its runtime in practice.

\textbf{Time complexity.} We analyze the cost in three parts: (1) the GNN Encoder, (2) intra-entity attention in the Query-Anchor Transformer, and (3) cross-entity attention:
(1) A standard message-passing layer costs $O(|\gE| \cdot D + N \cdot D^{2})$, comprising edge-wise aggregation and the node-wise update. The \uc carries the lifted axis as a batch dimension, so both terms are multiplied by the channel count, $F$ for the feature branch and $C$ for the label branch, with nothing else changed. (2) The Query-Anchor Transformer operates on the query and anchor sets rather than on the graph: intra-entity attention is self-attention over the $(F+C)$ channels of each entity and costs $O\big((|\gQ|+|\gA|)(F+C)^{2}D\big)$ per block. (3) Cross-entity attention lets each query attend to the anchors within each channel at $O\big(|\gQ||\gA|(F+C)D\big)$ per block. Letting $L_{\text{GNN}}$ denote the number of message-passing layers and $L_{\text{T}}$ the number of Transformer blocks, a forward pass costs
\[
O\big(L_{\text{GNN}}(|\gE| \cdot D + N \cdot D^{2})(F+C) \; + \; L_{\text{T}}\big[(|\gQ|+|\gA|)(F+C)^{2}D \; + \; |\gQ||\gA|(F+C)D\big]\big).
\]

Since $|\gQ|$ and $|\gA|$ are fixed at $256$ and $1024$, they are constants and can be absorbed, leaving
\[
O\big(L_{\text{GNN}}(|\gE| \cdot D + N \cdot D^{2})(F+C)+L_{\text{T}}(F+C)^2D\big).
\]

Against $O\big(L_{\text{GNN}}(|\gE| \cdot D + N \cdot D^{2})\big)$ for a message-passing network of the same depth, the \uc is roughly a factor $(F+C)$ more expensive, and scales with graph size in exactly the same way.

This is also a pessimistic estimate, since the \uc never performs a full-graph forward pass: each step samples anchors and queries and extracts $k$-hop subgraphs around them, so $|\gE|$ and $N$ above are those of the sampled subgraphs rather than of the dataset.

\looseness=-1
\textbf{Runtime.} With a single fixed checkpoint on a TPUv7, a complete pass over the test set takes under five minutes on \emph{roman-empire}, \emph{amazon-ratings} and \emph{arxiv}. A cost of that order per pass keeps training and repeated evaluation over many epochs tractable, in spite of the $(F+C)$ factor in the time complexity.

\section{Sampling, Label Masking, and Training Step}
\label{app:sampling}
This section gives the sampling procedure of \cref{subsec:data_handling} in full: the distribution the query and anchor sets are drawn from, the masking applied to query labels before they enter the model, and the resulting training step written out as pseudocode.

Let $\mathcal{V}_{tr}$ denote the set of elements in the training split, which are nodes for node- and edge-level tasks and graphs for graph-level tasks. At each training step we draw
\[
\gA \sim \mathrm{Unif}\big(\{S \subseteq \mathcal{V}_{tr} : |S| = n_A\}\big), \qquad
\gQ \mid \gA \sim \mathrm{Unif}\big(\{S \subseteq \mathcal{V}_{tr} \setminus \gA : |S| = n_Q\}\big)
\]
with $n_A = 1024$ and $n_Q = 256$, giving the joint law
\[
P(\gA, \gQ) = \binom{|\mathcal{V}_{tr}|}{n_A}^{-1}\binom{|\mathcal{V}_{tr}| - n_A}{n_Q}^{-1}.
\]
Drawing $\gQ$ from $\mathcal{V}_{tr} \setminus \gA$ is what enforces $\gQ \cap \gA = \emptyset$. The distribution is exchangeable over elements: there is no degree-, class- or centrality-based weighting, so every element is marginally equally likely to serve as an anchor or as a query. Label masking is then the deterministic map
\[
\widetilde\mY_v = \mY_v \ \text{ if } v \in \gA, \qquad \widetilde\mY_v = \mathbf{0} \ \text{ if } v \in \gQ,
\]
applied before $\phi_{\text{label}}$, so that no query label can enter message passing or attention. For link prediction, given an observed training edge $(u,v)$, the source $u$ is the query and $v$ a positive anchor, and $256$ negative anchors are drawn uniformly without replacement from the non-neighbors of $u$ in the training graph. At inference, $\gA$ is the training split, or a fixed subsample of it where the split exceeds the anchor budget, and $\gQ$ is the set of target elements.

\begin{algorithm}[htbp]
\caption{One \uc training step for node classification.}
\label{alg:uc_step}
\begin{algorithmic}[1]
\Require graph $\gG=(\gV,\gE,\mX,\mY)$; training split $\mathcal{V}_{tr}$; budgets $n_A,n_Q$; hops $k$; layers $L_{\text{GNN}}, L_{\text{T}}$
\Ensure  cross-entropy loss over the query set
\State $\gA \sim \mathrm{Unif}\big(\{S \subseteq \mathcal{V}_{tr} : |S| = n_A\}\big)$ \Comment{anchors}
\State $\gQ \sim \mathrm{Unif}\big(\{S \subseteq \mathcal{V}_{tr}\setminus\gA : |S| = n_Q\}\big)$ \Comment{queries; enforces $\gQ\cap\gA=\emptyset$}
\State $\widetilde\mY_v \gets \mY_v$ for $v\in\gA$;\quad $\widetilde\mY_v \gets \mathbf{0}$ for $v\in\gQ$ \Comment{label masking}
\State $\mathcal{G}_{\text{uni}} \gets \bigcup_{r\in\gQ\cup\gA}\mathcal{G}^k(r)$ \Comment{subgraph sampling}
\State $\mX \gets \textsc{ZNormalize}(\mX)$ \Comment{feature-wise}
\State $(\mX^{0})_{n,f,:} \gets \phi_{\text{feat}}(\mX_{n,f})\ \ \forall n,f$ \Comment{feature lifting}
\State $(\mY^{0})_{n,c,:} \gets \phi_{\text{label}}(\widetilde\mY_{n,c})\ \ \forall n,c$ \Comment{label lifting}
\For{$l = 0$ \textbf{to} $L_{\text{GNN}}-1$} \Comment{GNN Encoder}
    \State $\mX^{l+1} \gets \texttt{GNN}(\mX^{l},\mathcal{G}_{\text{uni}})$;\quad $\mY^{l+1} \gets \texttt{GNN}(\mY^{l},\mathcal{G}_{\text{uni}})$
\EndFor
\State retain root representations $\{\mX_r,\mY_r\}_{r\in\gQ\cup\gA}$
\For{$l = 1$ \textbf{to} $L_{\text{T}}$} \Comment{Query-Anchor Transformer}
    \ForAll{$r \in \gQ\cup\gA$} \Comment{intra-entity attention}
        \State $[\Bar\mX_r,\Bar\mY_r] \gets \texttt{TransformerBlock}([\mX_r \parallel \mY_r],[\mX_r \parallel \mY_r])$
    \EndFor
    \State $\mX_{\gQ} \gets \texttt{TransformerBlock}(\Bar\mX_{\gQ},\Bar\mX_{\gA})$ \Comment{cross-entity attention}
    \State $\mY_{\gQ} \gets \texttt{TransformerBlock}(\Bar\mY_{\gQ},\Bar\mY_{\gA})$
    \State $\mX_{\gA} \gets \Bar\mX_{\gA}$;\quad $\mY_{\gA} \gets \Bar\mY_{\gA}$ \Comment{anchors unchanged by queries}
\EndFor
\ForAll{$q\in\gQ,\ a\in\gA$} \Comment{Similarity Estimator}
    \State $\vs_{q,a} \gets \tfrac{1}{2D}\big(\tfrac{1}{F}\langle\mX_q,\mX_a\rangle_{\text{Frob}} + \tfrac{1}{C}\langle\mY_q,\mY_a\rangle_{\text{Frob}}\big)$
\EndFor
\State $\mM_{c,a} \gets \mathbbm{1}[\text{anchor } a \text{ belongs to class } c]$ \Comment{anchor-class mask}
\State $\vz_c \gets \operatorname{LME}(\vs,\mM)_c\ \ \forall c\in[C]$ \Comment{Anchor-Class Aggregator}
\State \Return $\mathrm{CrossEntropy}\big(\operatorname{softmax}(\vz),\, \mY_{\gQ}\big)$ \Comment{loss over queries only}
\end{algorithmic}
\end{algorithm}

\textbf{Training step.} The training step follows the description in \cref{subsec:data_handling,subsec:model_architecture}, and \cref{alg:uc_step} states the resulting procedure for node classification. Node regression replaces the one-hot labels with the soft bins of \cref{subsec:regression}; edge- and graph-level tasks change only which entities serve as queries and anchors, as described in \cref{subsec:link_graph_level}. Note that cross-entity attention updates the query representations only; anchors carry their intra-entity form forward unchanged, as recorded on the last line of the loop. The asymmetry is deliberate and follows the same convention as in-context learning over tables \citep{tabpfnv2,tabicl2025}. The anchor set is the context against which a prediction is made, and a context that absorbed information from the query would no longer be the same context for every query: an anchor's representation would depend on which query happened to be scored alongside it, so the prediction for one target would shift with the other targets in the batch. Keeping the anchors fixed removes that coupling, makes each query's prediction a function of the context alone, and allows the anchor representations to be computed once and reused across the whole query set.

The loop in \cref{alg:uc_step} over $r \in \gQ \cup \gA$ and the one over $q\in\gQ,\ a\in\gA$ are written elementwise for clarity only. Both are implemented as batched tensor operations, with the entity and channel axes carried as batch dimensions, so neither is materialized as an explicit loop at runtime.

\section{Datasets}
\label{app:datasets}
Our dataset selection is motivated by recent works calling for a paradigm shift in graph learning benchmarks. First, responding to the position paper \citep{bechlerspeicher2025positiongraphlearninglose}, we move beyond narrow academic domains to target impactful applications from e-commerce, crowdsourcing, and circuit design modeled in GraphLand and GraphBench. This prevents evaluation overfitting and stagnation by exposing the model to complex, cross-domain abstractions. Second, following the empirical analysis of \citet{platonov2023critical}, we discard popular but flawed benchmarks (like \textit{Squirrel} and \textit{Chameleon}) that suffer from data leakage caused by node duplication. Instead, we prioritize more reliable, diverse datasets that properly evaluate model behavior on both homophilic and heterophilic structures.
Third, graph-level benchmarks built to probe a specific property have not always turned out to require it: the Long Range Graph Benchmark was shown to close most of its reported gap once baselines are tuned and normalization is corrected, leaving little evidence that long-range interaction is what it measures \citep{tonshoff2024gap} and the widely used \textit{zinc} regression target is similarly narrow, being largely recoverable from cycle counts rather than from the molecular structure that quantum-chemical properties depend on \citep{dwivedi2022benchmarkinggraphneuralnetworks}. 

\textbf{GraphBench Electronic Circuits.} We therefore use the Electronic Circuits graph regression datasets of GraphBench \citep{graphbench} for our graph-level evaluation. 
We found, however, that the new splits in the repository are substantially different in size from those originally reported in GraphBench which makes reported baseline GNN performance incomparable. 
We implemented new baselines - classical ML baselines coming from the \texttt{sklearn} package which use \emph{49-dimensional} topological features: component counts (9 dims), number of nodes, edges, density, connected components, mean/max/std degree, top 15 eigenvalues of undirected simple graph, mnde feature mean (9 dims), node feature std (9 dims).

The \emph{Supervised GNN} is GIN with best hyperparameters found from running sweeps on EC datasets. The resulting architecture includes 4 layers with hidden dimension of 128 and was trained with Adam with learning rate 0.001.

\Cref{tab:datasets} provides summary statistics for the datasets. For the GraphLand datasets, we use both the Random Low (\emph{RL}) splits, which divide train, validation, and test nodes into $10\% / 10\% / 80\%$, and the Random High (\emph{RH}) splits, which divide them into $50\% / 25\% / 25\%$.

\begin{table}[htbp]
    \centering
    \caption{Summary statistics of the datasets used in our experiments. Node and edge counts are averaged per graph, and edges are counted after symmetrization, matching the graph representation used at runtime. The link-prediction datasets are reported separately in \cref{tab:lp_datasets}. EC graphs carry no node features, so we turn node types into one-hot encoded features. For the EC datasets, the number of graphs is given per train/validation/test split, and the EFF and VOUT targets share the same graphs.}
    \label{tab:datasets}

    \small
    \setlength{\tabcolsep}{5pt}
    \begin{tabular}{llrrrrr}
        \toprule
        \textbf{Task} & \textbf{Dataset} & \textbf{Graphs} & \textbf{Nodes} & \textbf{Edges} & \textbf{Features} & \textbf{Classes} \\
        \midrule
        \multirow{9}{*}{\makecell[l]{\textbf{Node}\\\textbf{Classification}}}
        & amazon-ratings  & 1 &     24,492 &     93,050 &       300 &        5 \\
        & artnet-exp      & 1 &     50,405 &    280,348 &        75 &        2 \\
        & arxiv           & 1 &    169,343 &  1,157,799 &       128 &       40 \\
        & city-reviews    & 1 &    148,801 &  1,165,415 &        37 &        2 \\
        & full-cora       & 1 &     19,793 &     63,421 &     8,710 &       70 \\
        & full-dblp       & 1 &     17,716 &     52,867 &     1,639 &        4 \\
        & hm-categories   & 1 &     46,563 & 10,730,995 &        35 &       21 \\
        & roman-empire    & 1 &     22,662 &     32,927 &       300 &       18 \\
        & tolokers-2      & 1 &     11,758 &    519,000 &        16 &        2 \\
        \midrule
        \multirow{4}{*}{\makecell[l]{\textbf{Node}\\\textbf{Regression}}}
        & artnet-views    & 1 &     50,405 &    280,348 &        50 &       -- \\
        & city-roads-L    & 1 &    142,257 &    231,550 &        26 &       -- \\
        & city-roads-M    & 1 &     57,073 &    107,104 &        26 &       -- \\
        & twitch-views    & 1 &    168,114 &  6,797,557 &         4 &       -- \\
        \midrule
        \multirow{3}{*}{\makecell[l]{\textbf{Graph}\\\textbf{Regression}}}
        & EC5  & 234,084/33,442/66,884 & 10.0 & 20.0 & 9 & -- \\
        & EC7  & 907/129/259 & 12.0 & 28.0 & 9 & -- \\
        & EC10  & 1,389/463/2,778 & 16.0 & 40.0 & 9 & -- \\
        \bottomrule
    \end{tabular}
\end{table}

The link prediction datasets were selected to cover sparse, dense, homophilic, heterophilic graphs. The datasets are listed apart from the rest because the protocol of \cref{sec:experiments} rebuilds each graph before use, so their counts differ from the node-level figures above; \textit{city-roads-M} differs most, as its upstream edges are directed. \Cref{tab:lp_datasets} reports them.

\textbf{In-depth link prediction protocol.} Every graph is made undirected, without self-loops; the node-level label is discarded and an edge-level link prediction task is constructed on the same graph. 
We hold out $1{,}000$ validation and $1{,}000$ test edges at random, using each source node at most once; the two sets share no edges or source nodes.
These $2{,}000$ edges are removed from the graph, so message passing uses only training edges and never sees a held-out edge.
We evaluate on a fixed subset of $200$ validation and $250$ test edges.
Each positive is ranked against $200$ random candidate targets, one shared list per split. We mask candidates that are the source or one of its true neighbors, and count ties as losses.

\begin{table}[htbp]
    \centering
    \caption{Statistics of the link prediction datasets after the preprocessing described above. \emph{Message-passing edges} excludes the $2{,}000$ held-out validation and test edges. Feature counts are post-transform (one-hot encoding and imputation), so \textit{city-roads-M} and \textit{tolokers-2} exceed the raw GraphLand counts.}
    \label{tab:lp_datasets}

    \small
    \begin{tabular}{lrrrrr}
        \toprule
        \textbf{Dataset} & \textbf{Nodes} & \textbf{Undirected edges} & \textbf{Message-passing edges} & \textbf{Features} & \textbf{Mean degree} \\
        \midrule
        roman-empire & 22,662 & 32,927  & 30,927  & 300 & 2.9  \\
        wiki-cs       & 11,701 & 215,603 & 213,603 & 300 & 36.9 \\
        city-roads-M & 57,073 & 107,104 & 105,104 & 68  & 3.8  \\
        tolokers-2   & 11,758 & 519,000 & 517,000 & 19  & 88.3 \\
        \bottomrule
    \end{tabular}
\end{table}

\section{Implementation Details and Hyperparameters}
\label{app:hyperparameters}
All results in \cref{sec:experiments} are obtained with a single pre-trained checkpoint, whose training and architectural hyperparameters are listed below.

\textbf{Training.} We pre-train the model for $20{,}000$ steps on TPUv7 accelerators using AdamW with a learning rate of 0.001, applying a dropout probability of 0.2. During training, the context anchor node set is sampled with a batch size of 1024, while the queried target nodes are sampled with a batch size of 256.

\textbf{Architecture.} Initial node embeddings are processed via a 2-layer MLP with a hidden dimension of 16. Our GNN Encoder is a 2-layer Graph Isomorphism Network (GIN) \citep{xu18} with a hidden dimension of 16, a self-loop weight initialization of $\epsilon = 0.1$, a maximum of 20 neighbors per hop, and interleaved Layer Normalization. Finally, the query-anchor module consists of 2 Transformer layers with a hidden dimension of 32. A logit soft-capping scalar of 50 is applied to stabilize gradient flows and prevent extreme logit magnitudes. We create 32 bins and centers for regression values. Then, each regression value has a probabilistic assignment to bins, derived using exponential softening.

\section{Limitations and Future Work}
\label{app:future}

While the \method (\uc) offers a unified foundation for graph learning, lifting features into 3D tensors for multiple queries and anchors introduces memory and inference overhead compared to standard task-specific GNNs. However, as demonstrated in our scaling ablations (\Cref{fig:anchor_scaling}), the \uc is highly sample-efficient. Utilizing only a small fraction ($1-5\%$) of available anchors recovers the vast majority of peak performance, effectively mitigating this computational bottleneck in practice.

A second limitation concerns task coverage. The evaluated checkpoint was pre-trained on node- and edge-level objectives only; although it already transfers zero-shot to some graph-level targets.

Looking forward, we intend to add graph-level datasets to the pre-training mixture, so that zero-shot transfer can be evaluated at that level as well, and to extend the \method to multi-relational domains; treating missing knowledge graph triplets or unlinked relational database records as queries evaluated against known relations could enable complex, multi-table inferences without requiring specialized heads.

\section{Detailed Experimental Results}
\label{app:detailed}
In this section, we present the full per-dataset results. Specifically, the results displayed in \Cref{fig:nc} can be found in \cref{tab:detailed_nc}, and in \cref{tab:detailed_nc_graphland} for the GraphLand datasets. The results displayed in \Cref{fig:regression} can be found in \cref{tab:detailed_nr} for node regression and \cref{tab:detailed_gr} for the graph regression dataset. The link prediction results displayed in \Cref{fig:link_prediction} can be found in \cref{tab:detailed_lp}.

\begin{table}[htbp]
    \centering
    \caption{Node classification accuracies.}
    \label{tab:detailed_nc}
    \resizebox{\textwidth}{!}{
    \begin{tabular}{l c c c c c}
        \toprule
        \textbf{Model} & \textbf{amazon-ratings} & \textbf{full-cora} & \textbf{full-dblp} & \textbf{roman-empire} & \textbf{arxiv} \\
        \midrule
        Supervised GNN & 47.18 $\pm$ 0.42 & 58.95 $\pm$ 0.55 & 73.87 $\pm$ 1.35 & 43.93 $\pm$ 0.45 & 73.65 $\pm$ 0.11 \\
        \midrule
        \multicolumn{6}{l}{\textit{Zero-Shot Transfer}} \\
        \midrule
        GraphAny & 42.80 $\pm$ 0.09 & 51.18 $\pm$ 0.78 & 71.48 $\pm$ 1.44 & 63.34 $\pm$ 0.58 & 58.85 $\pm$ 0.03 \\
        TS-Mean  & 42.27 $\pm$ 1.40 & 53.58 $\pm$ 0.73 & 66.42 $\pm$ 3.65 & 66.36 $\pm$ 1.02 & 56.33 $\pm$ 2.58 \\
        NodePFN  & 44.68 $\pm$ 0.48 & -      & 74.71 $\pm$ 0.39 & -      & -      \\
        UC       & 52.83 & 68.64 & 85.10 & 78.61 & 65.43 \\
        \midrule
        \multicolumn{6}{l}{\textit{Fine-Tuned}} \\
        \midrule
        UC       & 55.19 & 69.90 & 85.44 & 81.22 & 71.02 \\
        \bottomrule
    \end{tabular}
    }
\end{table}
\begin{table}[htbp]
    \centering
    \caption{Node classification accuracies on GraphLand datasets.}
    \label{tab:detailed_nc_graphland}
    \resizebox{\textwidth}{!}{
    \begin{tabular}{l cc cc cc cc}
        \toprule
        \textbf{Model}& \multicolumn{2}{c}{\textbf{hm-categories} (Acc)} & \multicolumn{2}{c}{\textbf{tolokers-2} (AP)} & \multicolumn{2}{c}{\textbf{city-reviews} (AP)} & \multicolumn{2}{c}{\textbf{artnet-exp} (AP)} \\
        \cmidrule(lr){2-3} \cmidrule(lr){4-5} \cmidrule(lr){6-7} \cmidrule(lr){8-9}
        & RL & RH & RL & RH & RL & RH & RL & RH \\
        \midrule
        Supervised GNN   & 67.96 $\pm$ 0.33 & 79.19 $\pm$ 0.21 & 53.78 $\pm$ 1.34 & 63.76 $\pm$ 1.30 & 77.67 $\pm$ 0.13 & 81.10 $\pm$ 0.11 & 46.62 $\pm$ 0.32 & 50.62 $\pm$ 0.35 \\
        \midrule
        \multicolumn{9}{l}{\textit{Zero-Shot Transfer}} \\
        \midrule
        TS-Mean   & 20.09 $\pm$ 1.29 & 15.48 $\pm$ 1.93 & 38.54 $\pm$ 0.94 & 31.83 $\pm$ 2.55 & 43.46 $\pm$ 5.17 & -      & 20.44 $\pm$ 1.05 & 13.38 $\pm$ 2.59 \\
        GraphPFN  & -      & -      & 60.33 $\pm$ 0.60 & - & 79.89 $\pm$ 0.10 & - & 50.85 $\pm$ 0.31 & - \\
        UC        & 48.69 & 54.83 & 67.18 & 69.26 & 86.53 & 86.57 & 65.46 & 66.92 \\
        \midrule
        \multicolumn{9}{l}{\textit{Fine-Tuned}} \\
        \midrule
        GraphPFN  & -      & -      & 60.90 $\pm$ 0.98 & -      & 80.49 $\pm$ 0.16 & -      & 50.64 $\pm$ 1.27 & -      \\
        UC        & 56.29 & 70.98 & 72.92 & 78.97 & 88.28 & 89.43 & 70.19 & 74.47 \\
        \bottomrule
    \end{tabular}
    }
\end{table}
\begin{table}[htbp]
    \centering
    \caption{Node regression $R^2$ on GraphLand datasets.}
    \label{tab:detailed_nr}
    \resizebox{\textwidth}{!}{
    \begin{tabular}{l cc cc cc cc}
        \toprule
        \multirow{2}{*}{\textbf{Model}} & \multicolumn{2}{c}{\textbf{city-roads-M}} & \multicolumn{2}{c}{\textbf{city-roads-L}} & \multicolumn{2}{c}{\textbf{twitch-views}} & \multicolumn{2}{c}{\textbf{artnet-views}} \\
        \cmidrule(lr){2-3} \cmidrule(lr){4-5} \cmidrule(lr){6-7} \cmidrule(lr){8-9}
        & RL & RH & RL & RH & RL & RH & RL & RH \\
        \midrule
        \multicolumn{9}{l}{\textit{End-to-end}} \\
        \midrule
        XGBoost   & 59.14 $\pm$ 0.11 & 70.93 $\pm$ 0.05 & 53.75 $\pm$ 0.07 & 64.62 $\pm$ 0.07 & 13.39 $\pm$ 0.00 & 13.34 $\pm$ 0.02 & 32.74 $\pm$ 0.04 & 36.83 $\pm$ 0.06 \\
        Supervised GNN       & 59.11 $\pm$ 0.20 & 70.53 $\pm$ 0.40 & 53.43 $\pm$ 0.20 & 66.03 $\pm$ 0.24 & 72.93 $\pm$ 0.17 & 76.06 $\pm$ 0.30 & 53.36 $\pm$ 0.78 & 59.01 $\pm$ 0.52 \\
        \midrule
        \multicolumn{9}{l}{\textit{Zero-Shot Transfer}} \\
        \midrule
        GraphPFN  & 63.55 $\pm$ 0.29 & -      & -      & -      & 73.27 $\pm$ 0.13 & -      & 61.77 $\pm$ 0.23 & -      \\
        UC        & 60.54 & 74.08 & 55.58 & 70.98 & 66.09 & 66.02 & 50.85 & 52.34 \\
        \midrule
        \multicolumn{9}{l}{\textit{Fine-Tuned}} \\
        \midrule
        GraphPFN  & 66.04 $\pm$ 0.41 & - & -      & -      & 78.17 $\pm$ 0.13 & - & 63.90 $\pm$ 0.12 & - \\
        UC        & 62.73 & 77.08 & 56.27 & 74.38 & 68.69 & 71.29 & 56.76 & 59.69 \\
        \bottomrule
    \end{tabular}
    }
\end{table}
\begin{table}[htbp]
    \centering
    \caption{Link prediction Hits@10 on the four link prediction datasets.}
    \label{tab:detailed_lp}
    \begin{tabular}{l cccc}
        \toprule
        \textbf{Model} & \textbf{roman-empire} & \textbf{wiki-cs} & \textbf{city-roads-M} & \textbf{tolokers-2} \\
        \midrule
        \multicolumn{5}{l}{\textit{End-to-end}} \\
        \midrule
        SEAL    & 87.33 $\pm$ 1.15  & 86.53 $\pm$ 1.97 & 99.47 $\pm$ 0.61 & 79.73 $\pm$ 0.61 \\
        NBFNet  & 88.00 $\pm$ 0.01  & 92.00 $\pm$ 0.40 & 99.20 $\pm$ 0.01 & 92.53 $\pm$ 0.92 \\
        BUDDY   & 58.40 $\pm$ 11.09 & 83.60 $\pm$ 4.51 & 91.47 $\pm$ 5.43 & 72.93 $\pm$ 1.22 \\
        \midrule
        \multicolumn{5}{l}{\textit{Zero-Shot Transfer}} \\
        \midrule
        UC      & 72.00 & 87.60 & 99.60  & 74.00 \\
        \midrule
        \multicolumn{5}{l}{\textit{Fine-Tuned}} \\
        \midrule
        UC      & 84.40 & 91.20 & 100.00 & 73.60 \\
        \bottomrule
    \end{tabular}
\end{table}

\begin{table}[t]
\centering
\caption{Graph Regression RSE on Electronic Circuits datasets.}
\label{tab:detailed_gr}
\setlength{\tabcolsep}{5pt}
\begin{tabular}{l cc cc cc}
\toprule
& \multicolumn{2}{c}{\textbf{EC-5}} & \multicolumn{2}{c}{\textbf{EC-7}} & \multicolumn{2}{c}{\textbf{EC-10}} \\
\cmidrule(lr){2-3} \cmidrule(lr){4-5} \cmidrule(lr){6-7}
\textbf{Method} & EFF & VOUT & EFF & VOUT & EFF & VOUT \\
\midrule
\multicolumn{7}{l}{\textit{End-to-end}} \\
\midrule
Ridge ($\alpha{=}1.0$)          & 0.760 & 0.760 & 0.970 & 0.897 & 0.893 & 0.948 \\
KNN ($k{=}10$)                  & 0.778 & 0.783 & 1.009 & 1.001 & 0.935 & 1.053 \\
Random Forest (200, $d{=}5$)    & 0.752 & 0.751 & 0.988 & 0.926 & 0.889 & 0.952 \\
GBDT (50, $d{=}3$)              & 0.720 & 0.724 & 0.989 & 0.890 & 0.877 & 0.977 \\
Supervised GNN                  & 0.840 & 0.856 & 0.912 & 0.937 & 0.831 & 0.926 \\
\midrule
\multicolumn{1}{l}{\textit{Zero-Shot Transfer}} \\
\midrule
\uc                      & 0.808 & 0.771 & 1.160 & 1.102 & 0.996 & 1.205 \\
\midrule
\multicolumn{1}{l}{\textit{Fine-tuned}} \\
\midrule
\uc                           & 0.527 & 0.558 & 0.909 & 0.842 & 0.672 & 0.806 \\
\bottomrule
\end{tabular}
\end{table}

\end{document}